\documentclass[10pt, a4paper]{article}
\usepackage{amsmath}
\usepackage[table]{xcolor}
\usepackage[final]{lrec2026} 
\usepackage{algorithm}
\usepackage{algpseudocode}
\usepackage{listings}
\usepackage{xcolor}
\usepackage{tabularx}
\usepackage{booktabs}
\usepackage{makecell}
\usepackage{caption}
\usepackage{enumitem}
\usepackage{arydshln}
\usepackage{amsmath,amssymb}
\usepackage{subcaption}
\usepackage{hyperref}
\usepackage[defaultcolor=magenta]{changes}

\setlist{nosep}
\usepackage{titlesec}
\titlespacing*{\section}{0pt}{5pt}{3pt}
\titlespacing*{\subsection}{0pt}{5pt}{3pt}

\lstdefinestyle{promptsmall}{
  basicstyle=\ttfamily\scriptsize,
  backgroundcolor=\color{gray!10},
  frame=single,
  rulecolor=\color{gray!50},
  breaklines=true,
  columns=fullflexible
}

\usepackage[most]{tcolorbox}
\newtcolorbox{promptbox}[2][]{%
  colback=blue!5!white,    
  colframe=blue!75!black,  
  fonttitle=\bfseries,     
  title=#2,                
  #1                       
}

\newcommand{\expandeddataset}{\textsc{Lunguage}\textsuperscript{++}}
\newcommand{\pathwayframework}{Pathway Expansion Framework}
\algnewcommand\algorithmicinput{\textbf{Input:}}
\algnewcommand\algorithmicoutput{\textbf{Output:}}
\algnewcommand\Input{\item[\algorithmicinput]}
\algnewcommand\Output{\item[\algorithmicoutput]}

\title{Modeling Clinical Uncertainty in Radiology Reports: \\from Explicit Uncertainty Markers to Implicit Reasoning Pathways}

\name{
Paloma Rabaey\textsuperscript{1}\textsuperscript{*}, Jong Hak Moon\textsuperscript{2}\textsuperscript{*}, Jung-Oh Lee\textsuperscript{3}, Min Gwan Kim\textsuperscript{4}, \\
\bfseries\large{Hangyul Yoon\textsuperscript{2}, Thomas Demeester\textsuperscript{1}, Edward Choi\textsuperscript{2}}}

\address{
\textsuperscript{1}Ghent University -- imec, \textsuperscript{2}KAIST, \textsuperscript{3}Mount Sinai Hospital, \textsuperscript{4}Seoul National University Hospital\\
paloma.rabaey@ugent.be, jhak.moon@kaist.ac.kr\\
\textsuperscript{*}Joint first authors.}

\abstract{
Radiology reports are invaluable for clinical decision-making and hold great potential for automated analysis when structured into machine-readable formats. These reports 
often contain uncertainty, which we categorize into two distinct types: (i) Explicit uncertainty reflects doubt about the presence or absence of findings, conveyed through \emph{hedging} phrases. These vary in meaning depending on the context, making rule-based systems insufficient to quantify the level of uncertainty for specific findings; (ii) Implicit uncertainty arises when radiologists omit parts of their reasoning, recording only key findings or diagnoses. Here, it is often unclear whether omitted findings are truly absent or simply unmentioned for brevity. 
We address these challenges with a two-part framework. We quantify \emph{explicit uncertainty} by creating an expert-validated, LLM-based \textit{reference ranking} of common hedging phrases, and mapping each finding to a probability value based on this reference. In addition, we model \emph{implicit uncertainty} through an expansion framework that systematically adds characteristic sub-findings derived from expert-defined diagnostic pathways for 14 common diagnoses. 
Using these methods, we release \expandeddataset{}, an expanded, uncertainty-aware version of the \textsc{Lunguage} benchmark of fine-grained structured radiology reports. This enriched resource enables uncertainty-aware image classification, faithful diagnostic reasoning, and new investigations into the clinical impact of diagnostic uncertainty. 
 \\ \newline \Keywords{Radiology report, Chest X-ray, Clinical uncertainty, Diagnostic reasoning, Large language model}}

\begin{document}

\maketitleabstract

\section{Motivation and Related Work} \label{sec:introduction}
\begin{figure}[t]
\begin{center}
\includegraphics[width=\columnwidth]{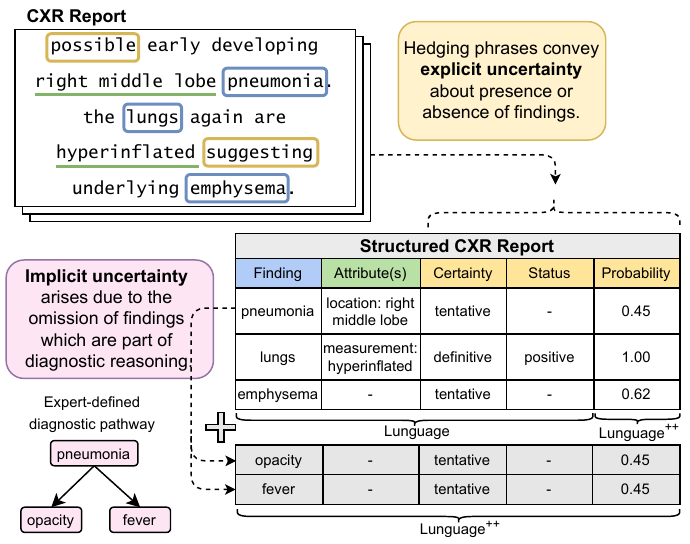}
\caption{\footnotesize \textbf{Two types of uncertainty in radiology reports that we address during structuring}, expanding the \textsc{Lunguage} dataset of structured CXR reports to form \expandeddataset{}. Explicit uncertainty is conveyed by hedging phrases that indicate tentative findings, whose (un)certainty we quantify with probabilities. Implicit uncertainty stems from findings that are not explicitly mentioned; we mitigate this by applying expert-defined diagnostic pathways to expand stated diagnoses with their characteristic sub-findings.}
\vspace{-2.2em}
\label{fig:uncertainty_radiology_reports}
\end{center}
\end{figure}

Radiology reports play a central role in clinical decision-making, serving as the primary medium through which radiologists communicate their interpretations and diagnostic impressions to referring physicians. 
These reports influence downstream diagnostic reasoning, and often determine treatment trajectories. 
As AI models have been increasingly used for both automated radiology report interpretation and generation, structuring frameworks have been developed to convert free-text reports into machine-readable formats that can be used for training and evaluation \citep{lunguage, chest_imagenome, radgraph, radgraph2}. 
A key challenge that needs to be addressed is that \textbf{radiology reports inherently contain uncertainty}. This uncertainty arises in
two distinct forms: explicit uncertainty, expressed directly through the language radiologists use to qualify their findings or diagnoses \citep{communicating_uncertainty, communication_doubt_radiology}, and implicit uncertainty, which emerges from the selective and often incomplete nature of what is recorded in the report \citep{linguistic_uncertainty_clinical}.
Figure \ref{fig:uncertainty_radiology_reports} shows how these two forms of uncertainty appear in chest X-ray (CXR) reports, which are especially prone to contain uncertainty \citep{uncertainty_nlp, chexpert}.

\textbf{Explicit uncertainty} arises when radiologists convey doubt \textbf{about the presence or absence of a finding or diagnosis}, typically through \emph{hedging} phrases such as ``probably'', ``possible'', ``suggesting'', or ``may represent'', among many others. Prior work examining these expressions has highlighted their prevalence, revealing the mention of uncertain diagnoses in a high proportion of CXR reports \citep{chexpert, lunguage}. 
The use of hedging in radiology reports is deliberate and meaningful: it enables radiologists to communicate diagnostic uncertainty that is crucial for appropriate clinical interpretation and decision-making \citep{quantitative_uncertainty, communicating_uncertainty}. Automated systems should similarly be designed to recognize and account for this uncertainty, just as human readers of radiology reports are trained to do.

Previous work has extracted uncertainty using rule-based systems \citep{chexpert, mimic_cxr_jpg, cad_chest} in ternary classification settings (positive / uncertain / negative). These approaches are limited in scope, using predefined vocabularies of hedging phrases that trigger an uncertain label when detected in the report. In this work, we go beyond discrete labeling by \textbf{quantifying uncertainty as a continuous probability between 0 and 1}, taking into account the specific hedging phrases and their sentence-level context (see Figure~\ref{fig:uncertainty_radiology_reports}).
Earlier attempts to assign probabilistic meanings to hedging expressions have typically relied on human judgments, asking experts to rate phrases or position them along a scale of certainty \citep{communication_doubt_radiology, diagnostic_uncertainty_radiology}. However, these approaches proved unreliable, as the interpretation of hedging varies widely across radiologists
. To address this, we introduce an automated approach that estimates the probability of a finding by leveraging large language models to perform pairwise comparisons of uncertainty expressions, constructing a relative ranking that is then mapped to a continuous probability.

Beyond overt language, reports also embody a subtler form of \textbf{implicit uncertainty} that has received little systematic attention. 
Radiologists frequently \textbf{omit portions of their diagnostic reasoning, documenting only key findings} or final impressions to maintain conciseness and ensure that the main message is easily understood by the reader \citep{ implicit_cognitive}. For example, a report may state ``congestive heart failure'' without mentioning sub-findings that are commonly associated with the condition, such as ``consolidation'' or ``cardiomegaly''. Consequently, it is often unclear whether unmentioned findings are truly absent or simply unrecorded, even though knowing this distinction is crucial, since misinterpreting unrecorded evidence as true absence can systematically bias the data and distort the inferred diagnostic reasoning \citep{implicit1, implicit_cognitive, implicit_bias, implicit_pitfall}. 

In other words, implicit uncertainty does not stem from linguistic ambiguity, but from the selective and incomplete nature of clinical reporting \citep{linguistic_uncertainty_clinical} -- what evidence is 
stated, abstracted, or left unstated within the report. 
Disentangling these possibilities requires contextual understanding of diagnostic logic and domain expertise, making implicit uncertainty difficult to model and largely unaddressed in prior research.
Understanding and modeling these uncertainties is not only critical for structuring radiology reports but also for developing AI systems that faithfully capture the reasoning process of radiologists. By explicitly representing both what is uncertain and what is implied, such structured resources enable more reliable training and evaluation of medical AI models in uncertainty-aware report generation and interpretation.

To address this gap, we aim to fill in the missing findings that are not explicitly mentioned but are implied by the stated diagnoses when structuring radiology reports, as is shown in Figure \ref{fig:uncertainty_radiology_reports}. We \textbf{construct expert-defined diagnostic pathways for 14 common CXR conditions}, capturing characteristic sub-findings typically observed with high likelihood (>80\%). These pathways are then used to enrich the original reports by deterministically adding sub-findings that support explicitly mentioned diagnoses, with diagnostic certainty (derived from our explicit uncertainty extraction pipeline) propagated to each added finding. This approach results in expanded structured reports that more accurately reflect the radiologist's underlying reasoning.

In summary, our contributions are as follows: 
\begin{itemize}
    \vspace{-.2em}
    \item \textbf{Quantifying explicit uncertainty}: We introduce a comprehensive framework to estimate the probability of findings in radiology reports, taking into account hedging phrases and their sentence-level context. As part of this framework, we publish an expert-validated reference ranking of common hedging phrases.
    \item \textbf{Addressing implicit uncertainty}: We present the first framework to model implicit uncertainty in radiology reports by releasing diagnostic pathways for 14 common CXR diagnoses and integrating them into a rule-based framework.
    This framework reconstructs omitted diagnostic evidence by inferring the sub-findings that support each diagnosis.
    \item \textbf{Releasing \expandeddataset{}}, which extends the \textsc{Lunguage} dataset of structured radiology reports \citep{lunguage} by incorporating our techniques for capturing explicit and implicit uncertainty.
\end{itemize}

Code can be found at our Github repository\footnote{\href{https://github.com/prabaey/lunguage_uncertainty}{github.com/prabaey/lunguage\_uncertainty}}, while \expandeddataset{} and related resources will be made available via Physionet.



\section{Lunguage Dataset} \label{sec:problem_setup}
We demonstrate our methods on the \textsc{Lunguage} dataset \citep{lunguage}, a benchmark dataset containing 1,473 annotated CXR reports from the MIMIC-CXR dataset \citep{mimic_cxr}. 
Each report has been structured into fine-grained \textit{(finding, relation, attribute)} triplets, where findings represent core clinical concepts (e.g. ``opacity'', ``pneumonia''), and relations specify their contextual links (e.g., “location”, “severity”) with corresponding attributes (e.g., “right lower lobe”, “moderate”). 
We use the \textsc{Lunguage} dataset because of its extensive granularity in the included findings, relations and attribute types.
Furthermore, each annotated finding includes a binary label (\textit{tentative} and \textit{definitive}) quantifying the confidence expressed by the radiologist -- hedging phrases (\textit{e.g.,} ``suggests'', ``cannot exclude'') lead to a \textit{tentative} label, while findings that lack such phrases are labeled \textit{definitive}.

In this work, we represent each report $\mathcal{R}$ as a structured set of $n_\mathcal{R}$ findings extracted from $m_\mathcal{R}$ sentences:
\begin{equation}
    \mathcal{R} = (\{(f_i, s_i, c_i, a_i) \}_{i=1}^{n_\mathcal{R}}, \{t_{j}\}_{j=1}^{m_\mathcal{R}}),
\label{eq:lunguage_entities}
\end{equation}
Here,  $t_j$ denotes the textual form of each \textbf{sentence}, $f_i$ indicates the \textbf{finding}, $s_i$ indicates the \textbf{status} (\textit{positive} if the finding is present and \textit{negative} if it is absent), $c_i$ indicates the \textbf{certainty} (\textit{definitive} or \textit{tentative}), and $a_i$ represents the \textbf{attributes} (e.g., location, morphology) associated with each finding. In \textsc{Lunguage}, there are 14,049 such structured findings. In our study, we exclude sentences from the \textit{history} section of the reports, as this section primarily contains information about patient symptoms or prior clinical records, whereas our analysis focuses on the \textit{findings} and \textit{impression} sections that convey diagnostic observations and reasoning.

\section{Explicit uncertainty} \label{sec:explicit_uncertainty}
\begin{figure*}[!t]
\begin{center}
\includegraphics[width=0.95\textwidth]{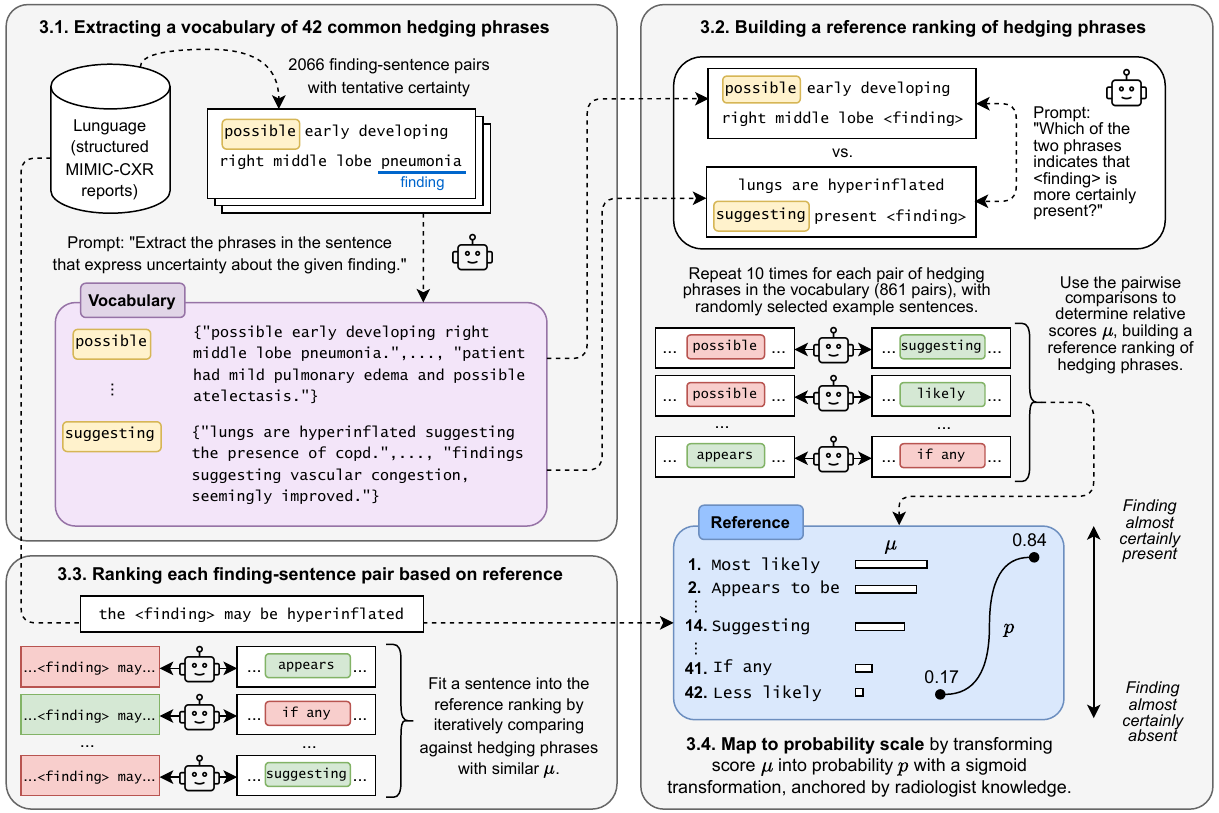}
\caption{\footnotesize \textbf{Strategy for assigning probabilities to finding-sentence pairs with tentative certainty in the \textsc{Lunguage} dataset}: We first build a vocabulary of common hedging phrases and the sentences in which these are used (Section~\ref{sec:explicit_step1}). Next, we leverage LLMs to construct a reference ranking of these phrases, by performing pairwise comparisons of examples sentences (Section~\ref{sec:explicit_step2}). Each finding-sentence pair is then compared against this reference (Section~\ref{sec:explicit_step3}) and is finally mapped to a probability (Section~\ref{sec:explicit_step4}). This approach ensures that the probability assigned to each finding reflects not only the hedging phrase itself but also the broader context in which it appears.
}
\label{fig:ranking_overview}
\end{center}
\vspace{-1.5em}
\end{figure*}
 
We assume we have a report $\mathcal{R}$, where the sentence $t_j$ expresses some uncertainty about finding $f_i$, in other words $c_i = tentative$. Across all reports in \textsc{Lunguage}, this leaves us with a subset of 2,066 tentative finding-sentence pairs. Our goal is to quantify this uncertainty by assigning a probability $p_i$ between 0 (finding certainly absent) and 1 (finding certainly present). In the remainder of this section, we ignore the explicit status $s_i$, as the target probability $p_i$ inherently captures the presence or absence of the finding. Furthermore, we disregard all findings where $c_i = definitive$, as there is no uncertainty in this case.
From this point onward, a \textit{sentence} refers to the complete text $t_j$, which contains a target \textit{finding} $f_i$ (e.g. ``pneumonia'') and one or more corresponding \textit{(hedging) phrase(s)} (e.g., ``possible''), expressing the degree of uncertainty associated with $f_i$.

Previous attempts to map hedging phrases to probabilities relied on expert ratings of individual phrases, an approach shown to be unreliable due to inconsistency across experts \citep{communication_doubt_radiology, diagnostic_uncertainty_radiology}. 
Furthermore, the context of the sentence beyond individual phrases should be taken into account: ``probably pneumonia'' conveys a different certainty than ``probably pneumonia given patient history'', even though both use the phrase ``probably''. 
To address these limitations, we adopt a different strategy, which relies on large language models (LLMs) to perform in-context pairwise comparisons between sentences conveying uncertainty. 
An overview of our framework is shown in Figure~\ref{fig:ranking_overview}. We now describe each step of the process in detail. Additional details can be found in Appendix \ref{app:explicit_uncertainty}.

\subsection{Extracting a vocabulary of common hedging phrases} \label{sec:explicit_step1}

For each finding-sentence pair, we automatically extract the hedging phrases that convey uncertainty about the finding $f_i$. We do this by prompting Gemini \citep{gemini}. 
The main part of the prompt is shown in Listing \ref{listing_extraction_prompt}, while the full prompt contains a system message, ten in-context examples and additional instructions (see Appendix \ref{app:vocab}). The prompt specifies the finding $f_i$ to avoid extraction of hedging phrases from the sentence which have nothing to do with that particular finding.
We include in our vocabulary all hedging phrases that were extracted ten times or more, resulting in a vocabulary of 42 hedging phrases. Each phrase is associated with a list of finding-sentence pairs where it was extracted. The five most common extracted hedging phrases include \textit{or} (373 times), \textit{likely} (239 times), \textit{may} (215 times), \textit{suggesting} (74 times), and \textit{cannot be excluded} (71 times); the full vocabulary is found in Appendix \ref{app:vocab}.

\begin{lstlisting}[style=promptsmall, caption={Prompt for hedging phrase extraction}, label={listing_extraction_prompt}]
Your task is to identify and extract only the words or phrases in the sentence that express uncertainty specifically about the given finding.
\end{lstlisting}

\subsection{Building a reference ranking of hedging phrases} \label{sec:explicit_step2}

We construct a reference ranking of the 42 common hedging phrases, where the top rank corresponds to a probability near 1 and the bottom rank near 0.

\paragraph{TrueSkill} 
We draw inspiration from ranking systems in competitive gaming, where player skill is inferred from the outcomes of matches.
Specifically, we employ the TrueSkill algorithm \citep{trueskill}, a Bayesian rating system that updates each item's mean skill level $\mu$ based on pairwise comparisons with other items. This way, TrueSkill efficiently infers a global ranking that reflects the relative skill level $\mu$ of each item. In our case, the items are hedging phrases.

\paragraph{LLM as a judge} To obtain a reliable and well-calibrated ranking, a large number of pairwise comparisons between phrases is required.
We therefore leverage LLMs to perform these comparisons automatically, employing three general-purpose LLMs (Gemini \citep{gemini}, GPT-4o \citep{gpt_4o}, and Claude \citep{claude}) and one domain-specific medical LLM (MedGemma \citep{medgemma}).\footnote{To deal with the sensitive nature of the reports in \textsc{Lunguage} we (i) ran a HIPAA-compliant GPT-4o model provided by Azure, (ii) revoked data retention rights for Gemini and Claude, and (iii) ran MedGemma locally.}
Since the preferred phrase can vary depending on the sentence context, we conduct ten comparisons for each pair of the 42 hedging phrases, where in each comparison, we randomly sample a sentence in which the phrase occurs.
We ask each LLM to identify which phrase conveys greater certainty about the presence of a finding, ensuring that such phrases will eventually receive a higher $\mu$. The main portion of the prompt is shown in Listing~\ref{listing_comparison_prompt}; the complete prompt additionally includes a system instruction, detailed task description, and three in-context examples (Appendix \ref{app:ref_ranking}). To ensure a more neutral comparison, the referenced finding $f_i$ within each sentence is masked as \textit{<finding>}.

\begin{lstlisting}[style=promptsmall, caption={Prompt for hedging phrase comparison}, label={listing_comparison_prompt}]
You will be given two sentences from radiology reports. Each sentence contains a placeholder <finding>, which represents a medical observation. Your task is to identify which sentence expresses a higher degree of certainty that the finding is present.
\end{lstlisting}

\begin{figure}[t]
\begin{center}
\includegraphics[width=\columnwidth]{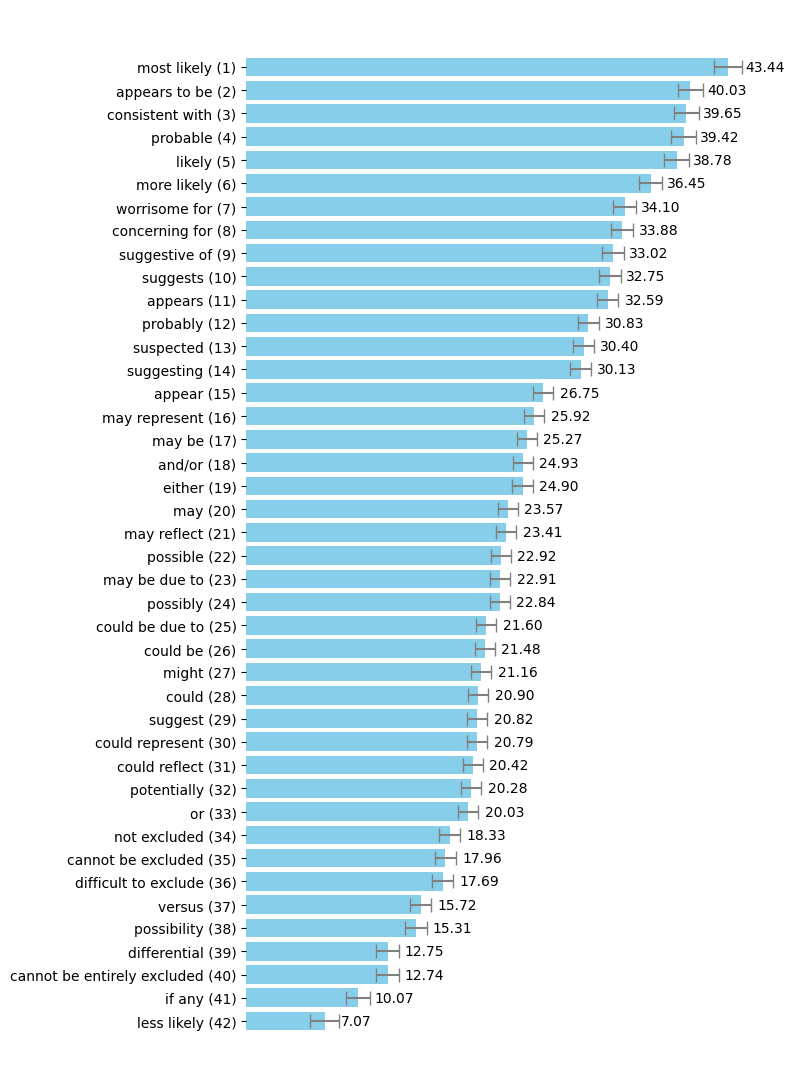}
\vspace{-2.em}
\caption{\textbf{Reference ranking of the 42 common hedging phrases in our vocabulary.} The mean skill level $\mu$ for each phrase is shown on the right, with the confidence $\sigma$ represented by the error bars. Phrases at the top of the ranking correspond to a high likelihood that the finding is present, while phrases at the bottom correspond to a high likelihood that the finding is absent.}
\label{fig:reference_ranking}
\end{center}
\vspace{-1.5em}
\end{figure}

This procedure yields $4 \times 8610$ pairwise comparisons, each treated as an independent comparison by the TrueSkill algorithm.
We execute TrueSkill 10 times with different random seeds that affect the order of matches, and then average the resulting $\mu$ values across runs.
The final averaged scores produce a stable reference ranking of hedging phrases, which is shown in Figure \ref{fig:reference_ranking}. Appendix \ref{app:ref_ranking} assesses the robustness of this reference ranking and explores inter-LLM agreement across the set of comparisons.

\begin{table}[t]
\centering
\small
\resizebox{\columnwidth}{!}{
\begin{tabular}{l c c c c c}
\hline
\textbf{Expert} & \textbf{Ref.} & \textbf{Gemini} & \textbf{GPT-4o} & \textbf{Claude} & \textbf{MedGem.} \\
\cmidrule(lr){1-2} \cmidrule(lr){3-6}
Radiologist & 0.80 & 0.80 & 0.82 & 0.72 & 0.84 \\
Radiologist & 0.80 & 0.76 & 0.82 & 0.76 & 0.80 \\
Internist & 0.86 & 0.86 & 0.88 & 0.82 & 0.90 \\
Oncologist & 0.72 & 0.68 & 0.74 & 0.76 & 0.72 \\
GP & 0.66 & 0.70 & 0.68 & 0.66 & 0.62 \\
GP & 0.82 & 0.82 & 0.84 & 0.78 & 0.82 \\
\cmidrule(lr){1-2} \cmidrule(lr){3-6}
Average & 0.78 & 0.77 & 0.80 & 0.75 & 0.77 \\
\end{tabular}
}
\caption{\textbf{Expert agreement with the reference ranking and individual LLMs.} We define agreement as the proportion of the 50 phrase pairs where expert and model judgments are concordant. GP = General Practitioner.}
\label{tab:expert_agreement}
\vspace{-1.5em}
\end{table}

\paragraph{Expert evaluation} To validate the final reference ranking, we conducted an expert evaluation study. We recruited two expert \textit{writers} (radiologists) and four expert \textit{readers} (internist, oncologist, and two general practitioners).
Each participant was presented with 50 pairs of hedging phrases, with five example sentences per phrase, all randomly sampled from our vocabulary. Participants were asked to select the phrase that conveyed a higher degree of certainty that the finding was present. All participants evaluated the same set of 50 phrase pairs. See Appendix \ref{app:ref_ranking} for additional details.

Table~\ref{tab:expert_agreement} presents the agreement between each expert and the reference ranking. Agreement is defined as the proportion of phrase pairs (out of 50) for which the expert's relative ordering of the two phrases matches that of the reference. 
In addition, we assess the agreement between each expert and each LLM. Since each LLM evaluated every phrase pair ten times using different example sentences, we first derive a consensus decision for each pair through majority voting, resolving ties at random. 

We see that the experts agree well with both the reference ranking and the individual LLMs. Furthermore, the inter-expert Fleiss' $\kappa$ is 0.72, indicating substantial agreement between experts. 
Among individual LLMs, GPT-4o shows the highest agreement with the experts, while Claude shows the lowest.
Despite Claude’s lower agreement on this limited subset of 50 pairs, we retain all LLMs in the reference ranking, to leverage the diversity of judgments across models, thereby improving the robustness of the TrueSkill-based ranking. Moreover, the expert evaluation covers only a small fraction of the full dataset and uses a different evaluation protocol than the sentence-level LLM comparisons, so occasional disagreements by a single LLM are not sufficient reason for exclusion. 

\subsection{Ranking each finding-sentence pair based on the reference ranking} \label{sec:explicit_step3}

With the reference ranking established, we fit each of the 2,066 tentative finding-sentence pairs from the \textsc{Lunguage} dataset into it. Note that we cannot simply use the hedging phrases extracted from the sentence to assign a rank directly, because (i) some phrases are too rare to appear in the reference vocabulary, and (ii) the context in which a phrase occurs can significantly alter its implied certainty. Here, we once again draw inspiration from competitive gaming: when a new player enters a game, TrueSkill quickly estimates their rank by identifying existing players with similar skill levels $\mu$. This is done by selecting opponents with the highest \textit{draw probability}, iteratively playing those games, and updating the skill level of the new player based on the outcome of each game \citep{trueskill}. 

To fit a finding–sentence pair into the reference ranking, we initialize the target sentence's parameter $\mu$ to the default value of $25$ and compute its draw probability against all opponent phrases in the reference ranking. The phrase with the highest draw probability is selected, and a corresponding sentence is randomly sampled from the vocabulary. Using the prompt from Listing~\ref{listing_comparison_prompt}, the target and opponent sentence are compared. During the first $K=10$ iterations, all four LLMs perform the comparison; thereafter, one LLM is selected at random to reduce cost. Based on the outcome, the TrueSkill algorithm updates $\mu$ for the target sentence. Draw probabilities are then recomputed, and the next opponent is selected accordingly, with each opponent limited to $N=5$ comparisons. The procedure terminates once the target sentence’s rank remains stable for ten consecutive steps, or after 100 iterations, whichever occurs first. Through hyperparameter tuning, we set $K$ to 10 and $N$ to 5. 
Applying our algorithm to the 2,066 tentative finding-sentence pairs in \textsc{Lunguage} incurred a total cost of \$92.16, averaging \$0.045/pair. The full algorithm, including experiments to validate our opponent selection strategy using \textit{draw probability}, can be found in Appendix \ref{app:fitting_ranking}.

\subsection{Map to probability scale} \label{sec:explicit_step4}

In the final step, we map the TrueSkill score $\mu$, which determines the position of each finding-sentence pair in the ranking, to a probability $p \in [0, 1]$.
This transformation is achieved using the sigmoid function $p = 1/(1+e^{-\alpha(\mu-\mu_0)})$,
where $\alpha$ controls the steepness of the curve and $\mu_0$ is the inflection point corresponding to a probability of 0.5. 

We determine $\alpha$ and $\mu_0$ using two anchor points: the desired probability $p_{bottom}$ for the phrase \textit{less likely} (lowest-ranked, $\mu = 7.07$), and $p_{top}$ for the phrase \textit{most likely} (highest-ranked, $\mu = 43.44$). To obtain these anchors, two radiologists independently reviewed ten example sentences per phrase and assigned a probability between 0 (``certainly absent'') and 1 (``certainly present''); see Appendix \ref{app:map_probability} for full instructions. Averaged across radiologists and examples, the resulting values were $p_{bottom} = 0.170$ ($95\%$ CI: $[0.156, 0.185]$) and $p_{top} = 0.839$ ($95\%$ CI: $[0.818, 0.860]$). These anchors yield $\alpha = 0.089$ and $\mu_0 = 24.89$.

After applying this mapping, the average probability for the 2,066 tentative finding-sentence pairs in \textsc{Lunguage} is 0.459, with a standard deviation of 0.185, a maximum of 0.892, and a minimum of 0.102. Table \ref{tab:disease_certainty} summarizes statistics for common CXR findings in the \textsc{Lunguage} dataset. For each finding, we report proportions of positive and negative cases with definitive certainty, and the mean, standard deviation, minimum, and maximum probability for cases with tentative certainty.

\begin{table*}[t]
\centering
\small
\renewcommand{\arraystretch}{1.2}
\resizebox{\textwidth}{!}{
\begin{tabular}{l c c c c c c c c c}
\hline
 & \multicolumn{2}{c}{\textbf{Certainty level}} & \multicolumn{2}{c}{\textbf{Definitive status}} & \multicolumn{4}{c}{\textbf{Tentative probability}} \\
\cmidrule(lr){2-3} \cmidrule(lr){4-5} \cmidrule(lr){6-9}
 \textbf{Finding}& \textbf{Definitive} & \textbf{Tentative} & \textbf{Positive} & \textbf{Negative} & \textbf{Avg.} & \textbf{Std.} & \textbf{Min.} & \textbf{Max.} \\
\hline
Pleural effusion & 1165 (86.2\%) & 186 (13.8\%) & 497 (42.7\%) & 668 (57.3\%) & 0.465 & 0.201 & 0.102 & 0.862 \\
Pneumothorax & 897 (98.8\%) & 11 (1.2\%) & 43 (4.8\%) & 854 (95.2\%) & 0.480 & 0.151 & 0.269 & 0.709 \\
Atelectasis & 419 (62.2\%) & 255 (37.8\%) & 408 (97.4\%) & 11 (2.6\%) & 0.536 & 0.156 & 0.145 & 0.876 \\
Pulmonary edema & 516 (83.5\%) & 102 (16.5\%) & 264 (51.3\%) & 252 (48.7\%) & 0.479 & 0.186 & 0.102 & 0.841 \\
Consolidation & 406 (84.1\%) & 77 (15.9\%) & 90 (22.2\%) & 316 (77.8\%) & 0.447 & 0.182 & 0.113 & 0.808 \\
Pneumonia & 225 (51.3\%) & 214 (48.7\%) & 32 (14.2\%) & 193 (85.8\%) & 0.423 & 0.221 & 0.102 & 0.872 \\
Cardiomegaly & 313 (97.8\%) & 7 (2.2\%) & 308 (98.4\%) & 5 (1.6\%) & 0.518 & 0.141 & 0.365 & 0.706 \\
Congestive heart failure & 49 (89.1\%) & 6 (10.9\%) & 23 (46.9\%) & 26 (53.1\%) & 0.591 & 0.248 & 0.102 & 0.765 \\
Emphysema & 40 (80.0\%) & 10 (20.0\%) & 40 (100.0\%) & 0 (0.0\%) & 0.518 & 0.134 & 0.314 & 0.660 \\
COPD & 18 (60.0\%) & 12 (40.0\%) & 18 (100.0\%) & 0 (0.0\%) & 0.537 & 0.050 & 0.453 & 0.594 \\
Fracture & 12 (85.7\%) & 2 (14.3\%) & 12 (100.0\%) & 0 (0.0\%) & 0.219 & 0.165 & 0.102 & 0.336 \\
Lung cancer & 6 (46.2\%) & 7 (53.8\%) & 6 (100.0\%) & 0 (0.0\%) & 0.531 & 0.171 & 0.249 & 0.662 \\
Tuberculosis & 3 (60.0\%) & 2 (40.0\%) & 2 (66.7\%) & 1 (33.3\%) & 0.413 & 0.051 & 0.377 & 0.449 \\
Bronchitis & 1 (50.0\%) & 1 (50\%) & 1 (100.0\%) & 0 (0.0\%) & 0.475 & 0.000 & 0.475 & 0.475 \\
\hline
\end{tabular}
}
\caption{\footnotesize \textbf{Certainty statistics for common CXR findings in the \textsc{Lunguage} dataset.} For each finding, we collect the instances in \textsc{Lunguage} where $f_i$ is mapped to the finding through its various lexical and contextual variants, following the approach described in Section~\ref{sec:expansion_system} (Finding Deduplication).}\label{tab:disease_certainty}
\end{table*}

\section{Implicit uncertainty} \label{sec:implicit_uncertainty}
\begin{figure*}[t]
\begin{center}
\includegraphics[width=\textwidth]{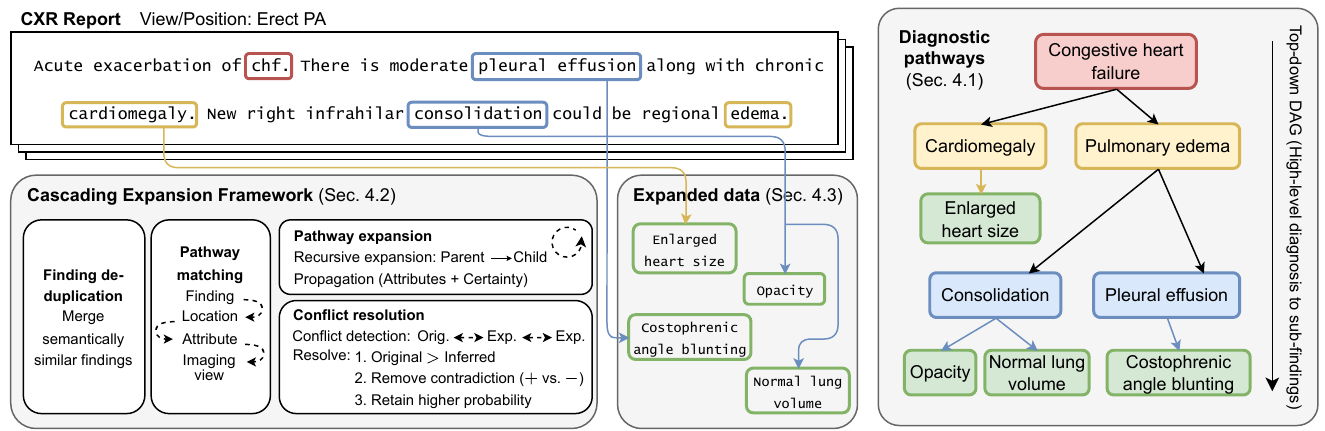}
\caption{\footnotesize \textbf{Overview of the \pathwayframework{}.}
The framework expands structured findings from \textsc{Lunguage} along diagnostic pathways (Section~\ref{sec:pathways}) to reconstruct omitted diagnostic evidence.
It comprises four stages—finding deduplication, pathway matching, pathway expansion, and conflict resolution (Section~\ref{sec:expansion_system})—that jointly ensure semantic coherence and clinical validity.
The resulting representation 
connects high-level diagnoses with their underlying evidence, forming \expandeddataset{}, which is further analyzed in Section~\ref{sec:analyis_of_expanded_dset}.}
\label{fig:pathway_overview}
\end{center}
\vspace{-1.5em}
\end{figure*}

Radiology reports often encode diagnostic reasoning implicitly \citep{implicit_bias, implicit_cognitive, implicit_pitfall, implicit1}, documenting high-level diagnoses while omitting intermediate supporting findings.
Although such abstraction enhances efficiency, it introduces ambiguity arising from unmentioned findings.
To resolve this, we propose a \textbf{\pathwayframework{}} that reconstructs omitted diagnostic evidence by systematically expanding structured findings along predefined \textbf{diagnostic pathways}.
This framework yields \expandeddataset{}, an expanded version of \textsc{Lunguage}, integrating explicit and implicit reasoning into a unified structured representation (Figure~\ref{fig:pathway_overview}).

\subsection{Diagnostic Pathway Construction}
\label{sec:pathways}

We formalize radiologists’ implicit diagnostic reasoning into explicit, machine-friendly formats that describe how each diagnosis decomposes into its characteristic radiographic findings. 
To preserve interpretability and clinical fidelity, we established three principles through expert consensus.
(i) \textbf{Exclusive mutual independence} ensures that each pathway differs from others by at least one defining piece of evidence.
(ii) \textbf{Specificity} ensures that observations directly contributing to diagnostic differentiation are included, emphasizing findings that are concrete and fine-grained.
(iii) \textbf{High-certainty features} retain only clinically consistent and reliable findings to minimize ambiguity and preserve clarity.
Pathways were then constructed through a two-stage expert-in-the-loop process grounded in these principles.
A radiologist and an AI researcher first defined subfindings for 14 common CXR diagnoses, referencing established radiological interpretation principles \citep{goodman2014felson, webb2011thoracic} to identify diagnostic criteria, imaging patterns, and hierarchical relations among findings.
These preliminary pathways were subsequently reviewed and refined through consensus by another radiologist and an oncologist.

Pathways are represented as Directed Acyclic Graphs (DAGs)~\citep{pearl2009causality}, capturing how higher-level diagnoses require lower-level manifestations within hierarchical reasoning. 
An example of a DAG corresponding to (part of) the diagnostic pathways for \textit{congestive heart failure} (\textit{CHF}), \textit{pulmonary edema}, \textit{pleural effusion}, and \textit{cardiomegaly} is shown in Figure \ref{fig:pathway_overview}.
Each node corresponds to an observable finding, and directed edges $(u, v)$ represent diagnostic relations $p_{path}(v|u)$ established through radiologist consensus, indicating that the sub-finding $v$ is typically observed when its parent $u$ is present, with likelihood greater than 0.8.
This top-down structure models the cascading sequence of radiographic evidence that underlies diagnostic inference.
Distinct pathways are defined for each diagnosis and further refined by imaging view and patient position, since these factors influence the findings one expects to observe. Table~\ref{tab:lunguage_pathway_expansion} reports statistical summaries of the diagnostic pathways, while the full set of pathways is provided in Appendix~\ref{app:pathway} and on our \href{https://github.com/prabaey/lunguage_uncertainty/blob/main/data_resources/dx_pathway.csv}{Github repository}.


\subsection{\pathwayframework{}}
\label{sec:expansion_system}

Building upon the diagnostic pathways, we design a \textit{\pathwayframework{}} that implements the top-down reasoning process of radiological interpretation to fill in missing sub-findings.  
Given structured findings from \textsc{Lunguage}~\citep{lunguage}, the framework reconstructs omitted intermediate findings by expanding each diagnosis along its predefined diagnostic pathway, following the recursive procedure detailed in Algorithm~\ref{alg:expand}.
It consists of four stages,
each of which ensure that reconstructed structures remain clinically faithful and logically consistent.

\paragraph{1. Finding Deduplication}
To ensure that subsequent reasoning operates on a consistent and non-redundant set of findings, overlapping or synonymous mentions referring to the same diagnostic concept at the same location (e.g., ``opacity in the basal segment of the lung'' and ``lung base opacity'') are identified and merged within each report.  
Pairwise cosine similarity is computed between clinical embeddings (BioLORD~\citep{remy2023biolord}) of all report findings, where each finding is linearized into a phrase combining its \textit{``entity, location, attributes''} following the \textsc{LunguageScore}~\citep{lunguage} formulation.  
Pairs exceeding a similarity threshold of 0.9 are merged into a single finding, while manually defined \textit{blacklist pairs} (e.g., left vs.~right) are excluded to prevent erroneous merges of semantically distinct findings. 
This precision-oriented threshold and manual blacklist correction minimize erroneous merges while ensuring that each finding is represented only once before expansion. The detailed blacklist pairs are provided in Appendix~\ref{app:blacklist}.

\paragraph{2. Pathway Matching}
After deduplication, the framework aligns each finding in \textsc{Lunguage} with its most appropriate diagnostic pathway~(Section~\ref{sec:pathways}) through a sequential process based on its \textit{finding}, \textit{location}, \textit{attributes}, and \textit{imaging view}. 
All terms are first normalized using the \textsc{Lunguage} vocabulary to ensure consistent terminology.  
The process begins with \textit{finding-level matching}, which identifies the core diagnostic concept but may yield ambiguous interpretations.  
For instance, ``fracture'' can denote either a medical device fracture (e.g., pacemaker lead) or a thoracic skeletal fracture (e.g., rib, spine).  
Such ambiguity is resolved through \textit{location matching}, which determines whether the finding pertains to an anatomical structure within the thorax or not, thereby selecting the appropriate pathway (e.g., device vs.~skeletal fracture).  
Subsequently, \textit{attribute matching} refines the alignment by specifying diagnostic stages such as ``acute'', ``chronic'', or ``healed''.  
Finally, \textit{imaging view} and \textit{patient orientation} (e.g., ``PA, erect'') define the view-specific pathway variant, ensuring that inferred findings remain anatomically observable within the imaging context. For the detailed process, see Appendix~\ref{appendix:ExpansionRules}.
\begin{table*}[t]
\centering
\scriptsize
\renewcommand{\arraystretch}{1.15}
\setlength{\tabcolsep}{5pt}
\resizebox{\linewidth}{!}{
\begin{tabular}{lcccc|rr}
\hline
 & \multicolumn{4}{c|}{\textbf{Diagnostic Pathway Structure}} 
 & \multicolumn{2}{c}{\textbf{Expansion within \textsc{Lunguage}$^\dagger$}} \\
\cmidrule(lr){2-5}\cmidrule(lr){6-7}
\textbf{Diagnosis} 
& \textbf{Variants} & \textbf{Pathways} & \textbf{Depth} & \textbf{Width} 
& \textbf{Expandable Finding} & \textbf{ Inferred Sub-findings} \\
\hline
Pleural effusion         & 2  &10  & 1 & 3 & 683 (4.9\%) & +915 (6.6\%) \\
Atelectasis              & 7  & 1  & 1 & 2 & 663 (4.7\%) & +1,215 (8.6\%) \\
Pulmonary edema          &12  & 2  & 2 & 4 & 366 (2.6\%) & +622 (4.4\%) \\
Consolidation            & 4  & 1  & 1 & 2 & 167 (1.2\%) & +801 (5.7\%) \\
Pneumonia                &22  & 3  & 2 & 3 & 249 (1.8\%) & +311 (2.2\%) \\
Cardiomegaly             & 1  & 1  & 1 & 1 & 315 (2.2\%) & +340 (2.4\%) \\
Congestive heart failure & 5  & 1  & 3  & 6 &  29 (0.2\%) & +79 (0.6\%) \\
Pneumothorax             & 5  & 2  & 1 & 4 &  54 (0.4\%) & +160 (1.1\%) \\
Emphysema                & 6  & 2  & 1 & 3 &  50 (0.4\%) & +98 (0.7\%) \\
COPD                     & 3  & 1  & 2 & 5 &  30 (0.2\%) & +88 (0.6\%) \\
Fracture                 &15  & 4  & 1  & 5 &  14 (0.1\%) & +25 (0.2\%) \\
Lung cancer              & 7  & 1  & 1  & 1 &  13 (0.1\%) & +12 (0.1\%) \\
Tuberculosis             & 5  & 2  & 1  & 3 &   4 (0.0\%) & +2 (0.0\%) \\
Bronchitis               & 4  & 2  & 1  & 3 &   2 (0.0\%) & +2 (0.0\%) \\
\hline
\rowcolor{gray!10}
\textbf{Total / Avg.}    
& \textbf{98} & \textbf{33} & \textbf{1.4} & \textbf{3.2}
& \textbf{2,639 (18.7\%)} & \textbf{+4,761 (33.9\%)} \\
\hline
\end{tabular}
}
\caption{
\textbf{Statistics of Diagnostic Pathways and \expandeddataset{}.} The left panel summarizes the hierarchical structure of 14 expert-defined diagnostic pathways (98 variants, 33 unique DAGs).  
The right panel quantifies their application within the 14,049 structured findings in \textsc{Lunguage}.  
A total of \textbf{2,636 findings (18.7\%)} were eligible for expansion (i.e., directly aligned with a pathway node) serving as anchors for hierarchical reasoning.  
From these anchors, the framework inferred \textbf{4,761 additional sub-findings (+33.9\%)}, representing previously implicit but clinically consistent observations reconstructed from the pathway hierarchy.  
\textit{$^\dagger$Percentages are relative to the 14,049 original findings in \textsc{Lunguage}. ``+” denotes additional findings inferred during expansion.}
}
\label{tab:lunguage_pathway_expansion}
\end{table*}

\paragraph{3. Pathway Expansion}

\begin{algorithm}[t]
\caption{Recursive Finding Expansion}
\label{alg:expand}
\small
\newcommand{\mycomment}[1]{\textcolor{gray}{\scriptsize\textit{#1}}}
\begin{algorithmic}[1]
\Input Parent node $f_i$ in pathway $\mathcal{P}$ with $s_i$, $c_i$, $p_i$, $a_i$
\Output Set $\mathcal{E}$ of all sub-findings expanded to leaf nodes

\Function{Expand}{$f_i, s_i, c_i, p_i, a_i, \mathcal{P}$}
  \State $\mathcal{E} \gets \emptyset$
  \If{$s_i \neq \textit{positive}$}
    \State \Return $\emptyset$ \mycomment{// Expand only positive findings}
  \EndIf
  
  \For{$f_j$ where $(f_i, f_j) \in \mathcal{P}$} \mycomment{// For each child $f_j$ in $\mathcal{P}$}
    \State $(s_j, c_j, p_j) \gets (s_i, c_i, p_i)$ \mycomment{// Inherit $f_i$ properties to $f_j$}
    \If{pathway $(f_i \!\rightarrow\! f_j)$ defines refinements}
      \State $a_j \gets a_i \cup \text{attributes specified for } f_j$
    \Else
      \State $a_j \gets a_i$ \mycomment{// Inherit as-is}
    \EndIf
    \State\mycomment{// Add expanded child $f_j$ to $\mathcal{E}$}
    \State $\mathcal{E} \gets \mathcal{E} \cup \{(f_j, s_j, c_j, p_j, a_j)\}$ 

    \If{$\mathrm{HasChildren}(f_j,\mathcal{P})$}
      \mycomment{// recurse for $f_j$}
      \State $\mathcal{E} \gets \mathcal{E}\; \cup$ \Call{Expand}{$f_j, s_j, c_j, p_j, a_j, \mathcal{P}$} 
       
    \EndIf
  \EndFor
  
  \State \Return $\mathcal{E}$ 
\EndFunction
\end{algorithmic}
\end{algorithm}
Building on the matched diagnostic pathways, the framework recursively expands structured findings in \textsc{Lunguage} by inferring omitted sub-findings through parent–child relations defined in each pathway (Algorithm~\ref{alg:expand}).
The expansion begins with a parent finding $f_i$, which corresponds to one of the 14 diagnoses in the diagnostic pathways $\mathcal{P}$. 
Each finding is represented by its diagnostic status $s_i$ (\textit{positive} or \textit{negative}), certainty $c_i$ (\textit{definite} or \textit{tentative}), probability $p_i$ (quantified in Section \ref{sec:explicit_uncertainty}), and attributes $a_i$ (e.g., location, morphology). 
During expansion, every child node in $\mathcal{P}$ inherits all diagnostic properties from its parent—$s_i$, $c_i$, $p_i$, and $a_i$—ensuring that each descendant finding reflects its parent’s context. 
For example, if $f_i$ is labeled as \textit{positive, tentative} with $p_i = 0.6$, all inferred sub-findings inherit the same label and probability. 
However, findings with $s_i = \textit{negative}$ are not expanded (line~3 in Algorithm~\ref{alg:expand}), since the pathway $\mathcal{P}$ is not reversible; for instance, ``no pneumonia'' does not imply ``no opacity.'' 
This inheritance (lines 7–11) reflects the propagation of diagnostic properties and the pathway-specific modification of attributes. Comprehensive expansion is achieved through recursive chaining (line~16): if a child finding $f_j$ corresponds to one of the 14 diagnoses in $\mathcal{P}$, it becomes a new parent node and triggers further expansion until reaching leaf nodes. For example, when ``pulmonary edema'' acts as a parent node, it expands to ``consolidation,'' which in turn recursively expands to its subfinding ``opacity.'' This recursive expansion ensures that all findings along the diagnostic pathway maintain consistent uncertainty estimates and coherent hierarchical relations.


\paragraph{4. Conflict Resolution}
Although pathway relations capture generally valid diagnostic dependencies (Section~\ref{sec:pathways}), 
clinical exceptions and inconsistent uncertain expression can lead to conflicting findings during expansion. 
These conflicts typically arise when multiple pathways generate overlapping findings, leading to discrepancies in diagnostic status (positive vs.~negative) or certainty (definitive vs.~tentative).  
Conflicts were classified as:  
(i) \textit{Original vs.~Expansion} — discrepancies between the original report and expansion (e.g., a resolved ``pulmonary edema'' reappearing from the CHF pathway); and  
(ii) \textit{Expansion vs.~Expansion} — contradictions among expansion (e.g., ``volume loss'' from atelectasis vs.~``normal volume'' from consolidation). To ensure clinical plausibility, all potential conflicts were reviewed with radiologist input and resolved through a rule-based consistency protocol. 
Resolution followed a sequential protocol:  
(1) original findings take precedence to preserve factual accuracy;  
(2) contradictory entries (e.g., positive vs.~negative) are removed; and  
(3) among uncertain cases, the instance with higher probability $p_i$ is retained.  
This process ensures logical coherence and clinical consistency across all expanded findings. The detailed procedure is presented in Appendix~\ref{app:conflict_analysis}.

\subsection{Analysis of \expandeddataset{}}
\label{sec:analyis_of_expanded_dset}
We analyze the diagnostic pathways constructed in Section~\ref{sec:pathways} and the expanded dataset generated by the framework in Section~\ref{sec:expansion_system}. 
This analysis quantifies the extent to which the framework reconstructs implicit diagnostic reasoning, propagates uncertainty, and maintains report-level coherence. Table~\ref{tab:lunguage_pathway_expansion} reports statistics for the full set of diagnostic pathways.

\paragraph{Diagnostic Pathway Diversity}
Pathway diversity arises from variations in \textit{imaging view}, \textit{finding}, \textit{location}, and \textit{attributes}, which inherently capture differences in disease stage (e.g., acute vs.~chronic).
Across \textbf{14 diagnostic categories} (e.g., pneumonia), we identified \textbf{98 disease variants} (e.g., hospital-acquired pneumonia) organized into \textbf{33 unique diagnostic pathways} (e.g., pneumonia vs.~lobar pneumonia), as shown in Table \ref{tab:lunguage_pathway_expansion}.
Each pathway is characterized by its \textbf{depth}, representing the maximum hierarchical expansion depth, and its \textbf{width}, denoting the number of leaf-level findings at the terminal layer. On average, pathways exhibit a depth of 1.4 and a width of 3.2, indicating that most reasoning chains consist of one to three inferential layers.  
Among them, \textit{pleural effusion} shows the greatest diversity (\textbf{10 pathways}) due to view-dependent fluid distribution, whereas \textit{congestive heart failure} exhibits the deepest hierarchy (depth 3), cascading through cardiomegaly, pulmonary edema, and pleural effusion (Figure \ref{fig:pathway_overview}).  
In contrast, simpler findings (e.g., \textit{cardiomegaly}) follow single-layer mappings (depth 1, width 1), directly linking diagnosis to observation.

\paragraph{Expansion Coverage}
As shown in Table \ref{tab:lunguage_pathway_expansion}, applying the \pathwayframework{} to 14,049 structured findings in \textsc{Lunguage} identified \textbf{2,639 findings (18.7\%)} that aligned with predefined diagnostic pathways, from which \textbf{4,761 additional sub-findings (+33.9\%)} were inferred.  
This shows that roughly one in five findings contained implicit diagnostic structure recoverable through pathways.  
Expansion was dominated by interdependent categories such as \textit{atelectasis} (\textbf{+1,215}), \textit{pleural effusion} (\textbf{+915}), and \textit{consolidation} (\textbf{+801}), which together accounted for more than half of all inferred findings.  
These conditions frequently co-occur or appear in hierarchical cascades naturally producing richer expansions, e.g.~effusion with edema or atelectasis. 


\paragraph{Conflict Resolution} To assess the stability of the expanded structures, we further examined conflicts that arose during expansion.  
Across the entire \textsc{Lunguage} dataset, such conflicts were rare, appearing in approximately \textbf{3.2\%} of cases overall:  
\textbf{0.9\%} arose between original and expanded findings (64.8\% status, 35.2\% certainty), and \textbf{2.3\%} between expansions (22.8\% status, 77.2\% certainty).
After conflict resolution, all remaining inconsistencies were resolved, confirming that the expanded dataset maintains both logical consistency and clinical validity.
Detailed analyses are presented in Appendix~\ref{app:conflict_analysis}.



\section{Conclusion} \label{sec:discussion} In this work, we present the first systematic approach to addressing both explicit and implicit uncertainty in radiology reports. We rigorously quantify the degree of \textbf{explicit uncertainty} of findings in CXR reports, through an LLM-based automated framework. 
While demonstrated on the \textsc{Lunguage} dataset, this framework can be applied to any CXR report corpus, enabling the enrichment of widely used benchmarks such as CheXpert \citep{chexpert} and MIMIC-CXR \citep{mimic_cxr_jpg} (of which \expandeddataset{} is only a subset) with continuous uncertainty measures.

In parallel, we expose and address \textbf{implicit uncertainty} in radiology reports arising from omitted elements of diagnostic reasoning. We introduce a rule-based expansion framework based on expert-defined diagnostic pathways for 14 common CXR diagnoses, which add characteristic sub-findings that may have been left unstated in the original reports. 
The diagnostic pathways can easily be reused for other tasks, while the expansion framework is applicable to any dataset that has been structured using the \textsc{Lunguage} framework proposed by \citet{lunguage}.

Together with our reusable frameworks, we release \expandeddataset{}, a benchmark dataset of structured radiology reports that includes continuous probabilities for all extracted findings and pathway-based expansions that expose previously omitted findings. This enriched resource supports a range of future research directions, including training uncertainty-aware CXR image classifiers, guiding vision-language models toward uncertainty-aware report interpretation and generation, and studying how diagnostic uncertainty influences downstream clinical outcomes.

\clearpage
\section*{Limitations}
For the \textbf{explicit uncertainty framework}, a key limitation lies in our reliance on LLM-based pairwise comparisons to construct the reference ranking of hedging phrases. While this approach offers the advantage of scalability (enabling 8,610 comparisons across four LLMs, far beyond what would be feasible with human raters) it also introduces dependency on model behavior. Moreover, our mapping from skill level $\mu$ to probability $p$ is based on assessments from only two radiologists; incorporating a larger expert pool would better capture the variability in how uncertainty is interpreted, as noted in prior studies \citep{communication_doubt_radiology, diagnostic_uncertainty_radiology}. Finally, the current framework incurs nontrivial costs when assigning probabilities to new sentences, due to repeated LLM comparisons with the reference ranking. Future work could mitigate this by training a lightweight, locally hosted model, fine-tuned on our high-quality LLM comparison data.

For the \textbf{implicit uncertainty framework}, the limitations include both structural and clinical aspects.  
First, the diagnostic pathways are constructed as a top-down Directed Acyclic Graph that maps high-level diagnoses to lower-level sub-findings.  
While this design captures hierarchical reasoning, it remains unidirectional and cannot represent bottom-up or cyclic dependencies that naturally arise in clinical reasoning -- such as when multiple findings interact or reinforce one another to revise a diagnosis.  
Future work could extend this structure into a bidirectional or dynamically learnable graph representation (e.g., a Bayesian network) that allows reasoning to flow in both directions, thereby capturing the iterative nature of diagnostic interpretation.

Second, when expanding findings along diagnostic pathways, we currently assign the same probability to all child nodes.
For instance, if \textit{congestive heart failure (CHF)} has a probability of 0.8, its inferred sub-findings—\textit{cardiomegaly} and \textit{pulmonary edema}—are each assigned the same value ($p = 0.8$), even though in practice, one may be more certain (e.g., $p=0.9$) while the other less so (e.g., $p=0.6$).
This simplification overlooks interdependence and uncertainty calibration among findings, which future work could address by modeling probabilistic propagation that accounts for relative diagnostic confidence.

Lastly, despite being defined through high-probability relations ($p(v|u) > 0.8$), the pathways inherently encompass clinical exceptions.  
Even strongly associated findings may not hold under atypical imaging conditions or in the presence of comorbidities, occasionally causing conflicts between pathway-inferred and explicitly reported findings.  
These exceptions reflect the inherent variability of radiological practice rather than modeling errors, but they underscore the need for incorporating image-level verification and cross-modal reasoning to refine the framework.  
Future iterations could leverage visual grounding to reconcile such exceptions, ensuring that inferred reasoning aligns more closely with true clinical evidence.

\section*{Data and Code Availability}

\expandeddataset{} and related resources will be made available via Physionet. Code can be found at our Github repository: \href{https://github.com/prabaey/lunguage_uncertainty}{github.com/prabaey/lunguage\_uncertainty}.

\section*{Ethics Statement}

This research made use of and expanded the \textsc{Lunguage} dataset \citep{lunguage}. This dataset is derived from MIMIC-CXR \citep{mimic_cxr}, a public dataset of chest radiographs and free-text radiology reports. As this dataset contains sensitive patient information, we will follow Physionet's guidelines by publishing \expandeddataset{} under the same agreement as the source data. Furthermore, to avoid passing sensitive patient data to public LLM APIs, we followed Physionet's recommendations by (i) running a HIPAA-compliant GPT-4o model provided by Azure, (ii) revoking data retention rights for Gemini and Claude, and (iii) running MedGemma locally. 

\section*{Acknowledgments}

Paloma Rabaey’s research is funded by the Research Foundation Flanders (FWO Vlaanderen) with grant number 1170124N. Additional FWO funding was provided in the form of a travel grant, with grant number V414025N. This research also received funding from the Flemish government under the “Onderzoeksprogramma Artificiële Intelligentie (AI) Vlaanderen” programme.

This work was partly supported by the Institute of Information \& Communications Technology Planning \& Evaluation (IITP) grants (No.RS-2019-II190075, No.RS-2025-02304967), the Korea Health Industry Development Institute (KHIDI) grant (No.RS-2025-02213750), and National Research Foundation of Korea (NRF) grant (NRF-2020H1D3A2A03100945), funded by the Korean government (MSIT, MOHW).

The authors would like to thank Géraldine Deberdt, Stefan Heytens and Johan Decruyenaere for participating in the expert evaluation study. Additionally, we are grateful to Stefan Heytens and Johan Decruyenaere for their input on the conceptual design of the expert evaluation study and sharing their view on uncertainty in medical reporting. \\

\newpage
\section*{Bibliographical References}\label{sec:reference}

\bibliographystyle{lrec2026-natbib}
\bibliography{reference}


\clearpage
\appendix
\renewcommand{\thetable}{\Alph{section}\arabic{table}}
\renewcommand{\thefigure}{\Alph{section}\arabic{figure}}
\renewcommand{\thelstlisting}{\Alph{section}\arabic{lstlisting}}
\renewcommand{\thealgorithm}{\Alph{section}\arabic{algorithm}}
\section*{Appendix} \label{sec:appendix}
\makeatletter
\@addtoreset{table}{section}
\@addtoreset{figure}{section}
\@addtoreset{lstlisting}{section}
\@addtoreset{algorithm}{section}
\makeatother

\section{Explicit Uncertainty} \label{app:explicit_uncertainty}

\subsection{Vocabulary of hedging phrases} \label{app:vocab}

The full prompt that was used to extract hedging phrases for each finding-sentence pair in Lunguage, is shown in Listing \ref{full_listing_extraction_prompt}. We use this to prompt the Gemini model \textit{gemini-2.5-flash} with the maximum number of output tokens set to 1100, where 1000 tokens were assigned as a thinking budget. \textit{Temperature} and \textit{top\_p} parameters were set to their default values of 1. The total price to extract all hedging phrases was around \$1.

\begin{lstlisting}[style=promptsmall, caption={Full prompt for hedging phrase extraction}, label={full_listing_extraction_prompt}]
SYSTEM: You are a radiologist who is given pairs of entities and sentences. Each entity appears in the corresponding sentence. Your task is to identify and extract only the words or phrases in the sentence that express uncertainty specifically about the given entity.

TASK: You are given pairs of entities and sentences. Each entity appears in the corresponding sentence. Your task is to identify and extract only the words or phrases in the sentence that express uncertainty specifically about the given entity.

Important notes:
- The sentence may mention multiple entities, but you should extract uncertainty clues only for the specified entity. Please refer to the examples below that show how you should handle such cases.
- Look for words or phrases that suggest uncertainty, speculation, approximation, possibility, or lack of definitiveness (e.g., "might", "possibly", "suggests", "appears to", "in some cases").
- Return a list of such uncertainty clues found in the sentence and relevant to the query entity.

Return your output as a list: ["<word or phrase 1>", "<word or phrase 2>", ...] 
If there are no uncertainty clues related to the given entity, return an empty list.

Below are 10 examples, after which you must complete the task for an unseen query. 

INPUT:
{
  "entity": "pulmonary edema",
  "sentence": "overall, however, there is a more focal airspace opacity in the left mid and lower lung, which may reflect asymmetric pulmonary edema or an infectious process, less likely atelectasis."
}
OUTPUT: 
["may", "or"]

INPUT:
{
  "entity": "atelectasis",
  "sentence": "overall, however, there is a more focal airspace opacity in the left mid and lower lung, which may reflect asymmetric pulmonary edema or an infectious process, less likely atelectasis."
}
OUTPUT: 
["less likely"]

... <redacted for brevity>

INPUT:
{
  "entity": "pleural effusion",
  "sentence": "a left pleural effusion and atelectasis obscure the left cardiac and hemidiaphragmatic contours more than the prior day."
}
OUTPUT: 
[]

Examine the entity and sentence pair below. If the sentence also talks about other entities, first identify the part of the sentence that is talking about the query entity, and then extract phrases expressing uncertainty specifically related to that entity.
Return your output as a list: ["<word or phrase 1>", "<word or phrase 2>", ...]. If there are no uncertainty clues related to the given entity, return an empty list.

INPUT: 
{
  "entity": {entity}
  "sentence": {sentence}
}
OUTPUT:

\end{lstlisting}

We manually reviewed and corrected all cases where more than one phrase was extracted (568 finding-sentence pairs), since these were prone to mistakes. Furthermore, we noticed that the LLM occasionally extracted phrases which did not relate to the uncertainty of the finding, but rather to other relational attributes which were already extracted in \textsc{Lunguage}: \textit{onset}, \textit{measurement} and \textit{severity}. Phrases that matched these attributes for the same finding-sentence pair were filtered out. The phrase "difficult to assess" was extracted in 11 cases, but was removed since it relates more to visual limitations rather than uncertainty.

\begin{figure}[t]
\begin{center}
\includegraphics[width=\columnwidth]{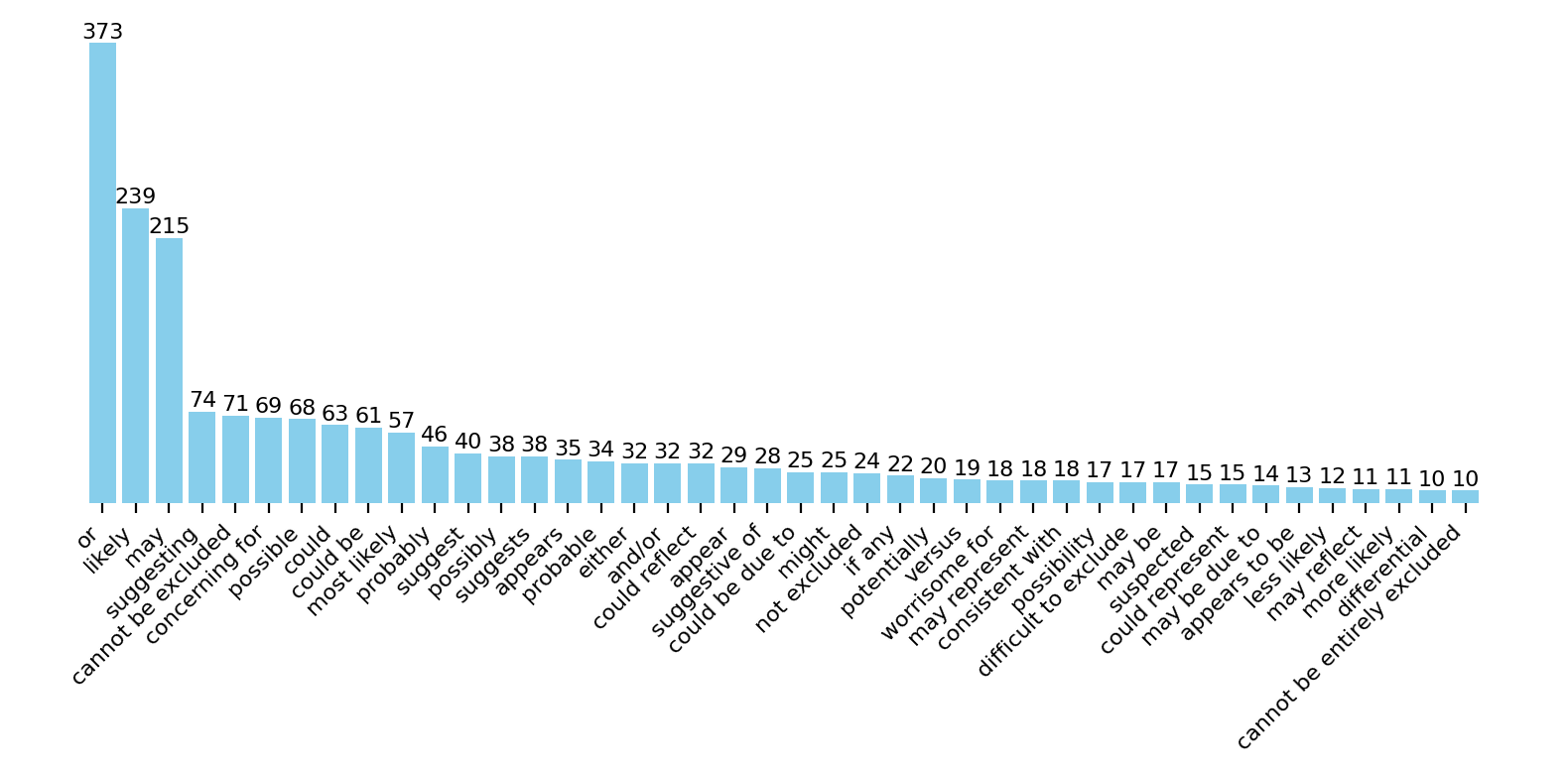}
\caption{\textbf{Our vocabulary of 42 common hedging phrases}, including how many times each phrase was extracted for tentative finding-sentence pairs in \textsc{Lunguage}.}
\label{fig:histogram_vocab}
\end{center}
\end{figure}

Figure \ref{fig:histogram_vocab} shows the number of times each of the 42 most common hedging phrases (i.e., occurring more than 10 times) were extracted from \textsc{Lunguage}.

\subsection{Reference ranking of hedging phrases} \label{app:ref_ranking}

\paragraph{TrueSkill details} The TrueSkill ranking system maintains a Bayesian belief in every item's skill by estimating two parameters: the average skill ($\mu$) and the degree of confidence in the skill ($\sigma$), together characterizing a Gaussian distribution $\mathcal{N}(\mu, \sigma)$. Initially, we set $\mu = 25$ and $\sigma = \frac{25}{3}^2$, which is the default. TrueSkill additionally defines a parameter $\beta^2$, which controls the variance of the item's performance around its average skill in a single match, set to $25/6$ per default. 

When we use an LLM to compare two hedging phrases, we are essentially playing a match where one hedging phrase is the winner. After each match, TrueSkill updates both $\mu$ and $\sigma$ using Bayesian inference: if an item performs better than expected, its $\mu$ increases, with the magnitude of this change depending on $\sigma$, and vice versa if it performs worse than expected. This mechanism is formalized in TrueSkill's update equations \citep{trueskill}, which ensure that $\sigma$ naturally shrinks as the system becomes more confident in its estimate, stabilizing each item's rating. Since the winner of a match is picked as the phrase that is closest to ``finding is certainly present'', a higher $\mu$ will be obtained for phrases closer to that part of the likelihood spectrum. In our case, we input all pairwise comparisons (across 861 pairs of hedging phrases, each repeated 10 times, with 4 LLMs) one after the other, in a random order. Since the final ranking is order-dependent, we repeat this 10 times with different random orders, averaging the final $\mu$ and $\sigma$ across runs.

\paragraph{Prompting details} The full prompt that was used to perform comparisons between pairs of hedging phrases is shown in Listing \ref{full_listing_comparison_prompt}. We executed this prompt for each pair of hedging phrases, repeating 10 times with different example sentences (\textit{sentence\_1} and \textit{sentence\_2}), and with each of the four LLMs: \textit{gemini-2.5-flash}, \textit{gpt-4o}, \textit{claude-sonnet-4}, \textit{medgemma-27b-text-it}. For each LLM, we set the maximum number of tokens to 100, \textit{temperature} to 1, \textit{top\_p} to 1. This time, we assigned zero thinking budget to Gemini, to pull all the models to the same level of reasoning capabilities. 

\begin{lstlisting}[style=promptsmall, caption={Full prompt for hedging phrase comparison}, label={full_listing_comparison_prompt}]
SYSTEM: You are a radiologist who is ranking sentences expressing uncertainty.

TASK: You will be given two sentences from radiology reports. Each sentence contains a placeholder "<finding>", which represents a medical observation (e.g., consolidation, effusion, nodule). Each sentence includes a phrase that expresses the degree of certainty about the presence or absence of the finding.

Assume there is a certainty spectrum ranging from:
    "<finding> is certainly absent"
    to
    "<finding> is certainly present"
Your task is to identify which sentence is **closer to "<finding> is certainly present"** on this scale, using the context of the sentence. In other words, your task is to identify which sentence expresses a higher degree of certainty that the finding is present.

Respond with **only** the chosen sentence (sentence_1 or sentence_2).

Here are some examples.

---

Example 1:
INPUT:
{
  "sentence_1": "interstitial markings are prominent, suggest possible mild <finding>.",
  "sentence_2": "allowing for low inspiratory volumes, the <finding> is probably unchanged."
}
OUTPUT:
"sentence_2"

---

Example 2:
INPUT:
{
  "sentence_1": "given the clinical presentation, <finding> must be suspected.",
  "sentence_2": "although this could represent severe <finding>, the possibility of supervening pneumonia or even developing ards must be considered."
}
OUTPUT:
"sentence_1"

---

Example 3:
INPUT:
{
  "sentence_1": "this could be either pneumonia in the left upper lobe or fissural <finding>.",
  "sentence_2": "the presence of a minimal left <finding> cannot be excluded, given blunting of the left costophrenic sinus.",
}
OUTPUT:
"sentence_1"

---

INPUT:
{
  "sentence_1": {sentence_1},
  "sentence_2": {sentence_2}
}

Which of the two sentences ("sentence_1" or "sentence_2") indicates that <finding> is more certainly present? Respond with your choice **only**.

OUTPUT:

\end{lstlisting}

\paragraph{Reference ranking} 

To assess the robustness of our reference ranking (shown in Figure \ref{fig:reference_ranking}), we constructed an alternative ranking using only five comparisons per phrase pair. Compared to the original ranking in Figure~\ref{fig:reference_ranking}, this reduced version showed an average absolute difference of 0.36 in $\mu$ and 0.76 in rank, with a Spearman correlation of 0.996 between both rankings. These results indicate that ten comparisons per pair provide a sufficiently stable and reliable ranking. We also compute pairwise agreement between the four LLMs across all 8610 comparisons, which is shown in Table \ref{tab:pairwise_agreement_LLMs}. The average agreement scores are 0.865 for Gemini, 0.861 for GPT-4o, 0.866 for Claude, and 0.866 for MedGemma. Fleiss' Kappa between all four models is 0.860 and Krippendorff's Alpha is 0.722. These values indicate strong but imperfect agreement across models, highlighting the need to integrate the judgments of all four LLMs when constructing the final ranking, as no single model can be assumed to provide the correct outcome in every case.

\begin{table}[htbp]
\centering
\caption{Pairwise agreement between LLMs across 8610 hedging phrase comparisons.}
\resizebox{\columnwidth}{!}{
\begin{tabular}{lcccc}
\hline
\textbf{Model} & \textbf{Gemini} & \textbf{GPT-4o} & \textbf{Claude} & \textbf{MedGemma} \\ 
\hline
\textbf{Gemini} & 1.000 & 0.861 & 0.868 & 0.866 \\
\textbf{GPT-4o} & 0.861 & 1.000 & 0.861 & 0.862 \\
\textbf{Claude} & 0.868 & 0.861 & 1.000 & 0.870 \\
\textbf{MedGemma} & 0.866 & 0.862 & 0.870 & 1.000 \\
\hline
\end{tabular}
}
\label{tab:pairwise_agreement_LLMs}
\end{table}

\begin{figure}[t]
\begin{center}
\includegraphics[width=\columnwidth]{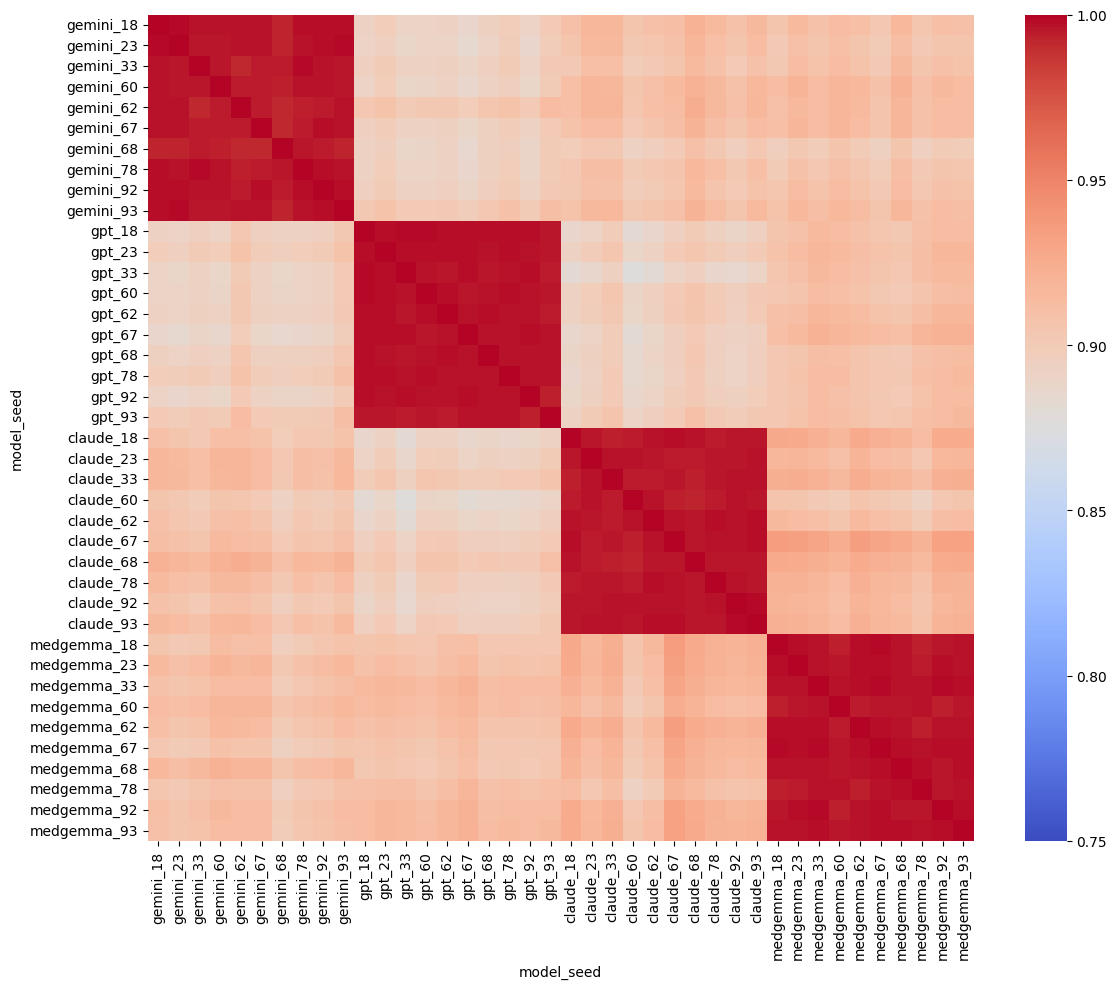}
\caption{\textbf{Pairwise Spearman rank correlations} between TrueSkill rankings obtained for individual LLMs, across 10 seeds.}
\label{fig:rank_correlations_LLMs}
\end{center}
\end{figure}

\begin{figure}[t]
\begin{center}
\includegraphics[width=\columnwidth]{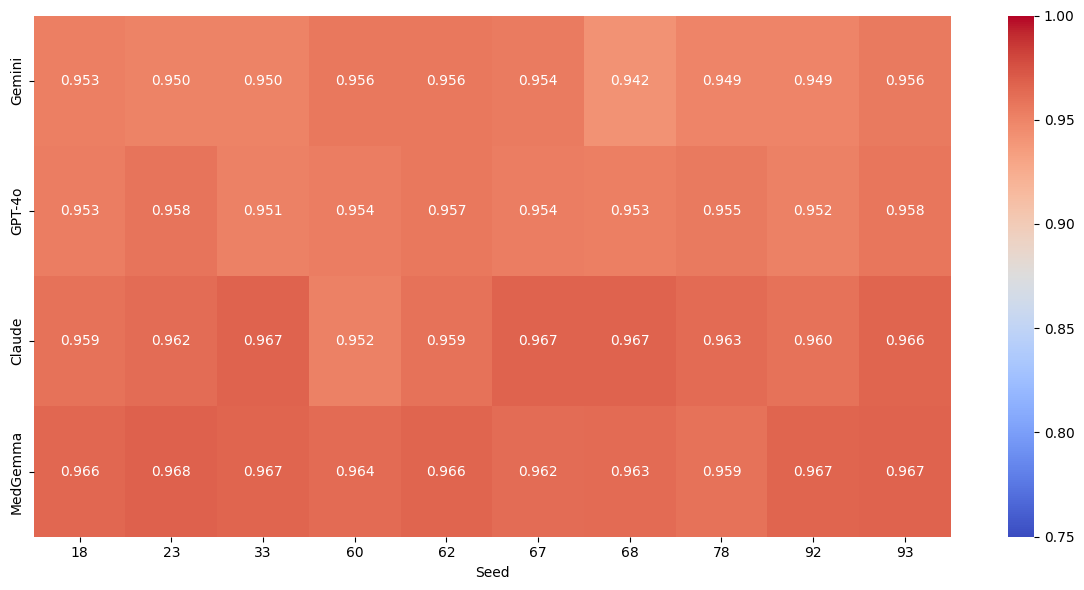}
\caption{\textbf{Spearman rank correlations with the reference ranking} for TrueSkill rankings obtained for individual LLMs, across 10 seeds.}
\label{fig:rank_correlations_LLMs_reference}
\end{center}
\end{figure}

To further explore the variation across LLMs, we apply the TrueSkill algorithm to each individual LLM's set of comparisons (8610 each), repeating 10 times for different random orderings of the matches. We then compute pairwise Spearman rank correlations between the obtained rankings. The full correlation plot is show in Figure \ref{fig:rank_correlations_LLMs}. Note that correlations are high for within-LLM comparisons, indicating that averaging across 10 ordering seeds should result in a stable ranking. Correlations are lower for across-LLM comparisons, once again indicating that relying on a single LLM to build our reference ranking would not be sufficient. As stated before, our final reference ranking (shown in Figure \ref{fig:reference_ranking}) is obtained by applying the TrueSkill algorithm to the full set of comparisons, repeating 10 times across ordering seeds and averaging the final $\mu$ and $\sigma$ across runs. Figure \ref{fig:rank_correlations_LLMs_reference} shows the Spearman rank correlation between each individual LLM's ranking across 10 seeds and the final reference ranking. Note that these correlations are generally higher than the pairwise correlations between individual LLMs from Figure \ref{fig:rank_correlations_LLMs}.

\begin{figure}[t]
\begin{center}
\includegraphics[width=\columnwidth]{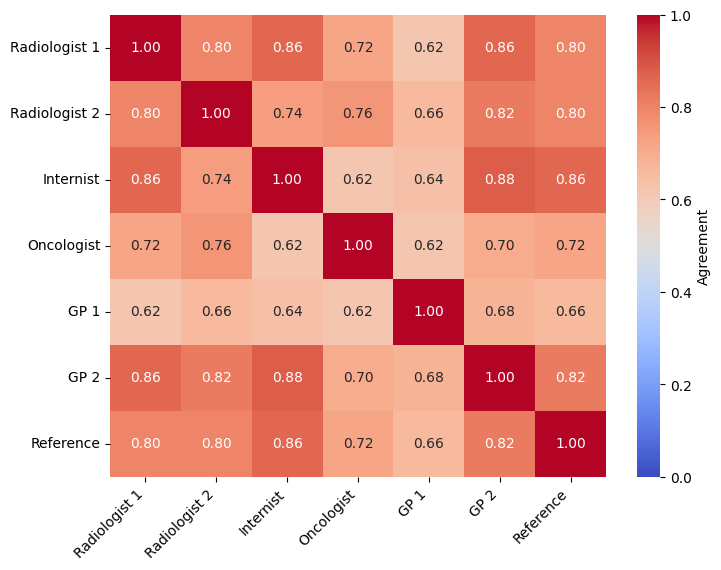}
\caption{\textbf{Pairwise agreement between experts and reference ranking}, across 50 hedging phrase pairs.}
\label{fig:pairwise_agreement_expert}
\end{center}
\end{figure}

\paragraph{Expert evaluation} Each participant was presented with 50 pairs of hedging phrases, with five example sentences per phrase. They were asked to choose the phrase that expresses a higher degree of certainty that the finding is present; an example is shown in Listing \ref{example_question}. The full survey, including detailed instructions for the evaluators, can be found on our \href{https://github.com/prabaey/lunguage_uncertainty/blob/main/data_resources/evaluation_study/survey_uncertainty.pdf}{Github repository}, in the file \texttt{survey\_uncertainty.pdf}. Table \ref{tab:expert_occupation} shows the occupation and experience of the six experts involved in our expert evaluation study. Figure \ref{fig:pairwise_agreement_expert} shows the pairwise agreement between experts, and between experts and the reference ranking. Agreement is defined as the proportion of phrase pairs (out of 50) for which the expert's relative ordering matches that of the other expert (or of the reference ranking).

\begin{lstlisting}[style=promptsmall, caption={Example question in expert evaluation study}, label={example_question}]
Phrase 1: difficult to exclude.
Example sentences:
- bilateral hilar vascular prominence is re-demonstrated with subtle <finding> in the left upper lung likely representing confluence of vasculature though a true nodule **difficult to exclude**.
- there is slight blunting of both costophrenic angles, felt most likely be due to overlying soft tissues, but a trace <finding> be **difficult to exclude**.
- no large <finding> is seen, although trace effusions are **difficult to exclude**.
- no large <finding> however, trace bilateral <finding>s **difficult to exclude**.
- mild <finding> is **difficult to exclude** in the correct clinical setting.

Phrase 2: appear.
Example sentences:
- the <finding> **appear** clear.
- the <finding> **appear** well inflated.
- the mediastinal and <finding> **appear** unchanged, allowing for differences in technique.
- mid <finding> **appear** intact.
- <finding> **appear** grossly intact.
\end{lstlisting}

\begin{table}[!ht]
\centering
\small
\renewcommand{\arraystretch}{1.2}
\begin{tabular}{l l l}
\hline
\textbf{Type} & \textbf{Occupation} & \textbf{Experience} \\
\hline
Writer & Radiologist & 5 to 10 years \\
Writer & Radiologist & 0 to 5 years \\
Reader & Internist & 30 to 40 years \\
Reader & Oncologist & 5 to 10 years \\
Reader & GP & 30 to 40 years \\
Reader & GP & 0 to 5 years \\
\hline
\end{tabular}
\caption{\textbf{Expert occupation and experience}}
\label{tab:expert_occupation}
\end{table}

\subsection{Fitting each finding-sentence pair into the reference ranking} \label{app:fitting_ranking}

\paragraph{Algorithm} Using Algorithm \ref{alg:fit_pair}, we can fit any tentative finding-sentence pair into the reference ranking. Here, the draw probability between the target sentence $t_{tar}$ with $(\mu_{tar}, \sigma_{tar})$, and an opponent phrase $opp$ with $(\mu_{opp}, \sigma_{opp})$ is calculated as follows: $DrawProb(t_{tar}, opp) = exp({\frac{-(\mu_{tar} - \mu_{opp})^2}{2c^2}})\sqrt{d}$, with $c^2 = 2\beta^2+\sigma_{tar}^2+\sigma_{opp}^2$, $d = 2\frac{\beta^2}{c^2}$, and $\beta^2 = 25/6$ \citep{trueskill}.

\begin{algorithm}[t]
\caption{\small Determine TrueSkill score $\mu$ for finding-sentence pair $(f_{tar}, t_{tar})$ based on reference ranking}
\label{alg:fit_pair}
\small
\begin{algorithmic}[1]
\State \textbf{Input:} Target sentence $t_{tar}$, reference ranking $\mathcal{R}$, LLMs, $K$, $N$, $\text{max\_steps}=100$, $\text{patience}=10$
\State \textbf{Output:} TrueSkill score $(\mu_{tar}, \sigma_{tar})$ for $t_{tar}$
\State Initialize $\mu_{tar} \gets 25$, $\sigma_{tar} \gets \frac{25}{3}^2$, $step \gets 0$, $s \gets 0$
\State Initialize $opp\_counts[\text{phrase}] \gets 0$, $\forall$ phrase $\in \mathcal{R}$
\While{$step < \text{max\_steps}$ \textbf{and} $s < \text{patience}$}
    \State Select $opp \in \mathcal{R}$ with $max$ $DrawProb(t_{tar}, opp)$
    \State \textbf{if} $opp\_counts[opp] \ge N$ \textbf{then} skip
    \State $opp\_counts[opp] \gets opp\_counts[opp] + 1$
    \State Randomly select sentence $t_{opp}$ containing $opp$ 
    \If{$step < K$} 
    \State $models \gets$ all LLMs
    \Else 
    \State $models \gets$ randomly pick 1 LLM
    \EndIf
    \State Compare $t_{tar}$ vs $t_{opp}$ using $models$
    \State Update TrueSkill $(\mu_{tar}, \sigma_{tar})$ for $t_{tar}$
    \State Recalculate rank $r_{tar}$ of $\mu_{tar}$ among $\mathcal{R}$
    \State \textbf{If} $r_{tar}$ unchanged \textbf{then} $s \gets s + 1$ \textbf{else} $s \gets 0$
    \State $step \gets step + 1$
\EndWhile
\end{algorithmic}
\end{algorithm}

\paragraph{Validation}

To validate our algorithm and the use of draw probability as an opponent selection strategy, we designed the following experiment. While the true rank of individual sentences in our dataset is unknown, we can assess performance using the hedging phrases in our reference vocabulary, whose ranks are known. Specifically, we simulate a modified reference ranking in which one of the 42 hedging phrases is temporarily excluded, then fit that phrase back into the ranking using a modified version of Algorithm \ref{alg:fit_pair}. We consider two opponent selection strategies: (i) the \textit{draw probability} strategy described previously, and (ii) a \textit{random} strategy, where an opponent is chosen uniformly at random at each step. We also test multiple LLM configurations: one using \textit{all} LLMs for the first $K$ iterations (as in Algorithm \ref{alg:fit_pair}) and then picking an LLM randomly, and another using a single LLM (\textit{Gemini}, \textit{GPT-4o}, \textit{Claude}, or \textit{MedGemma}) for all comparisons. Each configuration is repeated with 10 random seeds, which control sentence sampling, LLM selection, and opponent choice in the random setting. At each iteration, we compute the absolute distance between the current estimated rank of the phrase and its true rank in the reference ranking. If the algorithm converges before 100 iterations, the final rank is used for all subsequent iterations. We then average the absolute rank distances across all 42 phrases and 10 seeds. In these experiments, K was set to 10 and N to 5.

\begin{figure}[t]
\begin{center}
\includegraphics[width=\columnwidth]{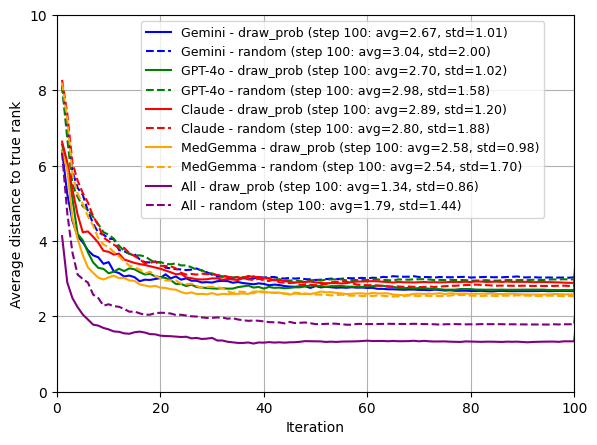}
\caption{\textbf{Average distance to the true rank across iterations of variants of Algorithm \ref{alg:fit_pair}}. We test different variants of the opponent selection strategy--\textit{draw probability} vs. \textit{random}--and LLMs used for comparisons--\textit{all} or \textit{single-LLM} (Gemini, GPT-4o, Claude, or MedGemma).}
\label{fig:ranking_experiments}
\end{center}
\end{figure}

Figure \ref{fig:ranking_experiments} presents the resulting performance across all configurations. Note that the strategy which uses \textit{draw probability} and \textit{all} LLMs performs best, which is indeed the strategy implemented in Algorithm \ref{alg:fit_pair}. While the \textit{random} strategy performs similarly to \textit{draw probability} in the \textit{single-LLM} setting, it leads to less stable runs, as evidenced by the standard deviation of the distance to the true rank at step 100, averaged across all phrases, which is shown in Figure \ref{fig:ranking_experiments} as well.

\begin{figure}[t]
    \centering
    \begin{subfigure}[b]{0.4\textwidth}
        \centering
        \includegraphics[width=\linewidth]{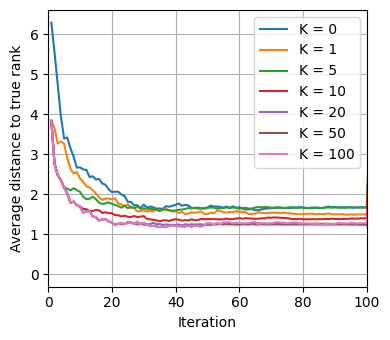}
        \caption{\textbf{$K$}}
        \label{fig:hyperparam_K}
    \end{subfigure}
    \hfill
    \begin{subfigure}[b]{0.4\textwidth}
        \centering
        \includegraphics[width=\linewidth]{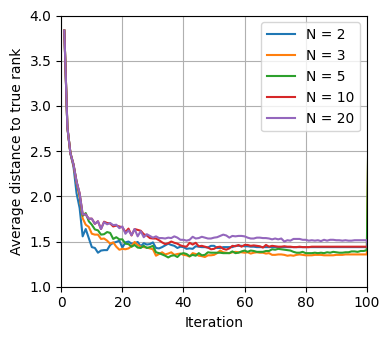}
        \caption{\textbf{$N$}}
        \label{fig:hyperparam_N}
    \end{subfigure}
    \caption{Hyperparameter tuning results for $K$ and $N$.}
    \label{fig:hyperparam_combined}
\end{figure}

\paragraph{Hyperparameters} Algorithm \ref{alg:fit_pair} contains hyperparameter $K$, which controls the number of iterations for which all four LLMs perform the phrase-sentence comparison, and hyperparameter $N$, which decides the number of times each opponent can be selected for comparison. Following the same experiment setup outlined above, with the \textit{draw probability} strategy and including \textit{all} LLMs, we perform hyperparameter optimization across the hedging phrases in our vocabulary. For $K$, we ran Algorithm \ref{alg:fit_pair} for $K \in [0, 1, 5, 10, 20, 50, 100]$ with $N = 5$. Figure \ref{fig:hyperparam_K} shows the average distance to the true rank in the reference ranking, averaged across all phrases. Setting $K = 10$ forms the right balance between cost-efficiency and performance. We performed the same experiment for $N \in [2, 3, 5, 10, 20]$, with $K = 10$, ultimately choosing $N = 5$  (Figure \ref{fig:hyperparam_N}). 

\paragraph{Results and cost} Applying Algorithm \ref{alg:fit_pair} to the 2,066 tentative finding-sentence pairs in \textsc{Lunguage} incurred a total cost of \$92.16, averaging \$0.045/pair. Specifically, we spent \$5.59 on Gemini, \$39.49 on GPT-4o, and \$45.12 on Claude, while MedGemma incurred no cost as it was run locally. On average, 24.61 ranking steps were required per finding-sentence pair, ranging from 10 to 100. The average $\mu$ value is 22.62, with a minimum of 0.41 and a maximum of 48.58, compared to a minimum of 7.07 and a maximum of 43.44 in the reference ranking (see Figure \ref{fig:reference_ranking}). The average rank when fit into the reference ranking is 23.87, with a minimum rank of 1 and a maximum rank of 43. 

\subsection{Map to probability scale} \label{app:map_probability}

\begin{figure}[t]
\begin{center}
\includegraphics[width=\columnwidth]{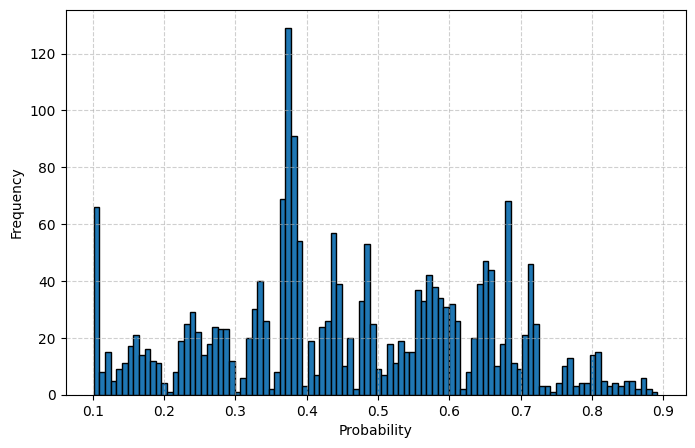}
\caption{\textbf{Histogram of probabilities in the tentative portion of \expandeddataset{}.}}
\label{fig:histgram_prob}
\end{center}
\end{figure}

We use an expert-informed sigmoid mapping (described in Section \ref{sec:explicit_step4}) to transform each $\mu$ into its probability value $p$. A histogram of the probabilities in the tentative portion of \expandeddataset{} is shown in Figure \ref{fig:histgram_prob}.

This mapping is anchored by the expert-defined probabilities for the phrases \textit{less likely} and \textit{most likely}. Two radiologists received the following instruction: 

\begin{lstlisting}[style=promptsmall]
On a scale of certainly absent (0) to certainly present (100), where would you place the phrase <phrase>, as it relates to the <finding> in each of the following sentences? Keep in mind that your answer may differ based on the context of the sentence.
\end{lstlisting}

They were asked to assign such probabilities for 10 example sentences, for both phrases. An example sentence for \textit{less likely} includes: ``Minimal blunting of the right costophrenic sulcus is more suggestive of similar slight atelectatic change, less likely persistent trace <finding>''. An example sentence for \textit{most likely} includes: ``Streaky predominantly right-sided mid and lower lung opacities are seen, most likely due to <finding>''. The full instructions are included in the survey, which can be found on our \href{https://github.com/prabaey/lunguage_uncertainty/blob/main/data_resources/evaluation_study/survey_uncertainty.pdf}{Github repository}, in the file \texttt{survey\_uncertainty.pdf}.

\makeatletter
\@addtoreset{table}{section}
\@addtoreset{figure}{section}
\@addtoreset{lstlisting}{section}
\@addtoreset{algorithm}{section}
\makeatother
\clearpage
\section{Implicit Uncertainty}
\label{appendix:implict_uncertainty}

\subsection{Diagnostic Pathways}
\label{app:pathway}
We present the diagnostic pathways defined in \textit{\pathwayframework{}}. In practice, the pathways encode context primarily along two axes: (i) \textbf{imaging view and patient positioning}, which determine which radiographic signs can be expected or meaningfully interpreted (e.g., pleural effusion on an erect PA view typically presents with costophrenic angle blunting, whereas on a supine AP view it more often appears as diffuse haziness over the pleural space); and (ii) \textbf{clinical acuity together with morphologic/anatomic subtype}, which jointly modulate how a diagnosis presents (e.g., pneumothorax shows peripheral pleural air with absent pulmonary lung markings, whereas tension pneumothorax additionally shows mediastinal shift; pleural effusion may be non-loculated or loculated). These conditioning factors are encoded in the pathway dictionary. During expansion, the \textit{\pathwayframework{}} uses these pathways to
recover omitted intermediate findings in a report. Table~\ref{tab:pathway_summary} and Figures~\ref{fig:pathway_panels_part1_final8} to \ref{fig:pathway_panels_part2_final6} present representative pathway variants for the 14 diagnoses. In the figures, each panel depicts the diagnosis as the root and its subfindings as children, with arrows indicating expert-defined dependencies. For example, in congestive heart failure the pathway links the root to cardiomegaly, pulmonary edema, and dyspnea (congestive heart failure $\rightarrow$ cardiomegaly + pulmonary edema + dyspnea). A dictionary with all pathway variants is also released on our 
\href{https://github.com/prabaey/lunguage_uncertainty/blob/main/data_resources/dx_pathway.csv}{Github repository}, in the file \texttt{dx\_pathway.csv}.

\tikzset{
    dagnode/.style={
        rectangle,
        draw=black,
        rounded corners,
        align=center,
        inner sep=2pt,
        font=\scriptsize,
        text width=1.8cm,
        minimum height=0.55cm
    }
}

\begin{figure*}[tbp]
\centering
\begin{minipage}[t]{0.45\linewidth}
\textbf{Pleural Effusion}\\[0.3em]
\begin{tikzpicture}[node distance=0.7cm]
\node[dagnode] (eff) {Pleural Effusion};

\node[dagnode, below=1cm of eff, xshift=-2.3cm] (cpa) {Costophrenic angle blunting (dp)};
\node[dagnode, below=1cm of eff, xshift=+2.3cm] (sup) {Diffuse/hazy opacity + layering fluid (dp)};

\node[dagnode, below=2.0cm of eff, xshift=0cm] (locu) {Loculated pleural opacity (dp)};

\draw[->] (eff) -- node[midway,sloped,above,font=\scriptsize]{erect} (cpa);
\draw[->] (eff) -- node[midway,sloped,above,font=\scriptsize]{supine} (sup);
\draw[->] (eff) -- node[midway,sloped,above,font=\scriptsize]{loculated} (locu);
\end{tikzpicture}
\end{minipage}
\hfill
\begin{minipage}[t]{0.45\linewidth}
\textbf{Pneumothorax}\\[0.3em]
\begin{tikzpicture}[node distance=0.7cm]
\node[dagnode] (ptx) {Pneumothorax};

\node[dagnode, below=1cm of ptx, xshift=-2.6cm] (air) {Pleural space lucency (dp) + pulmonary markings (dn) + periphery air (dp)};
\node[dagnode, below=1cm of ptx, xshift=+2.6cm] (shift) {Pleural space lucency (dp) + pulmonary markings (dn) + Mediastinal shift + large periphery air (dp)};

\draw[->] (ptx) -- (air);
\draw[->] (ptx) -- node[midway,sloped,above,font=\scriptsize]{tension} (shift);
\end{tikzpicture}
\end{minipage}

\vspace{0.9em}

\begin{minipage}[t]{0.45\linewidth}
\textbf{Consolidation}\\[0.3em]
\begin{tikzpicture}[node distance=0.7cm]
\node[dagnode] (cons) {Consolidation};

\node[dagnode, xshift=+3.3cm] (airspace) {Airspace opacity + volume loss (dn)};

\draw[->] (cons) -- (airspace);
\end{tikzpicture}
\end{minipage}
\hfill
\begin{minipage}[t]{0.45\linewidth}
\textbf{Atelectasis}\\[0.3em]
\begin{tikzpicture}[node distance=0.7cm]
\node[dagnode] (atel) {Atelectasis};

\node[dagnode, xshift=+3.3cm] (voloss) {Parenchymal opacity + Volume loss (dp)};

\draw[->] (atel) -- (voloss);
\end{tikzpicture}
\end{minipage}

\vspace{0.9em}

\begin{minipage}[t]{0.45\linewidth}
\textbf{Pneumonia}\\[0.3em]
\begin{tikzpicture}[node distance=0.7cm]
\node[dagnode] (pna) {Pneumonia};

\node[dagnode, below=1cm of pna, xshift=-2.6cm] (opac) {Parenchymal opacity + fever (dp)};
\node[dagnode, below=1cm of pna, xshift=+2.6cm] (fever) {consolidation + fever (dp)};


\draw[->] (pna) -- (opac);
\draw[->] (pna) -- node[midway,sloped,above,font=\scriptsize]{lobar} (fever);
\end{tikzpicture}
\end{minipage}
\hfill
\begin{minipage}[t]{0.45\linewidth}
\textbf{Pulmonary Edema}\\[0.3em]
\begin{tikzpicture}[node distance=0.7cm]
\node[dagnode] (p_edema) {Pulmonary Edema};

\node[dagnode, below=1cm of p_edema, xshift=-2.6cm] (septal) {interstitial opacity (dp)};

\node[dagnode, below=1cm of p_edema, xshift=+2.6cm] (fluffy) {consolidation + pleural effusion (dp)};

\draw[->] (p_edema) -- node[midway,sloped,above,font=\scriptsize]{interstitial} (septal);
\draw[->] (p_edema) -- node[midway,sloped,above,font=\scriptsize]{}(fluffy);
\end{tikzpicture}
\end{minipage}

\vspace{0.9em}

\begin{minipage}[t]{0.45\linewidth}
\textbf{Bronchitis}\\[0.3em]
\begin{tikzpicture}[node distance=0.7cm]
\node[dagnode] (bronch) {Bronchitis};

\node[dagnode, below=1cm of bronch, xshift=-2.6cm] (pbi) {Peribronchial opacity (dp)};
\node[dagnode, below=1cm of bronch, xshift=+2.6cm] (cprod) {Bronchial wall thickening + Chronic cough (dp)};


\draw[->] (bronch) -- (pbi);
\draw[->] (bronch) -- node[midway,sloped, above,font=\scriptsize]{chronic} (cprod);
\end{tikzpicture}
\end{minipage}
\hfill
\begin{minipage}[t]{0.45\linewidth}
\textbf{Cardiomegaly}\\[0.3em]
\begin{tikzpicture}[node distance=0.7cm]
\node[dagnode] (cardio) {Cardiomegaly};

\node[dagnode, xshift=+3.3cm] (heart) {Enlarged cardiac silhouette (dp)};

\draw[->] (cardio) -- (heart);
\end{tikzpicture}
\end{minipage}

\caption{
\textbf{Diagnostic pathway panels} (Part 1 of 2) illustrating exemplar pathways from the hand-crafted dictionary used by the pathway matching algorithm. Each panel is a small directed graph whose root node is a normalized diagnosis (\textit{finding}) and whose children are normalized sub-finding patterns over location, attributes, view, and (optionally) clinical context (e.g., patient symptom). Edges are annotated with the attribute or view modifiers that the matcher uses when checking joint compatibility between an input tuple \((\textit{finding}, \textit{loc}, \textit{attr}, \textit{view})\) and a pathway variant. Here, \textit{dp} stands for definitive positive, while \textit{dn} stands for definitive negative. Shown pathways correspond to: Pleural Effusion, Pneumothorax, Consolidation, Atelectasis, Pneumonia, Pulmonary Edema, Bronchitis, and Cardiomegaly. For Pleural Effusion, the branches encode erect versus supine presentations and a loculated variant. For Pneumothorax, the variants capture simple pneumothorax, tension pneumothorax, and hydropneumothorax. For Pneumonia and Pulmonary Edema, the pathways encode typical parenchymal and interstitial/alveolar patterns (consolidation) together with key clinical or co-occurring radiographic findings (e.g., fever, pleural effusion). Bronchitis is represented as an airway-centered process, while Consolidation, Atelectasis, and Cardiomegaly map directly to their canonical imaging signatures. All pathway structure shown here is taken directly from the released pathway dictionary and is the same structure searched by \texttt{BestPathwayMatch}.}
\label{fig:pathway_panels_part1_final8}
\end{figure*}
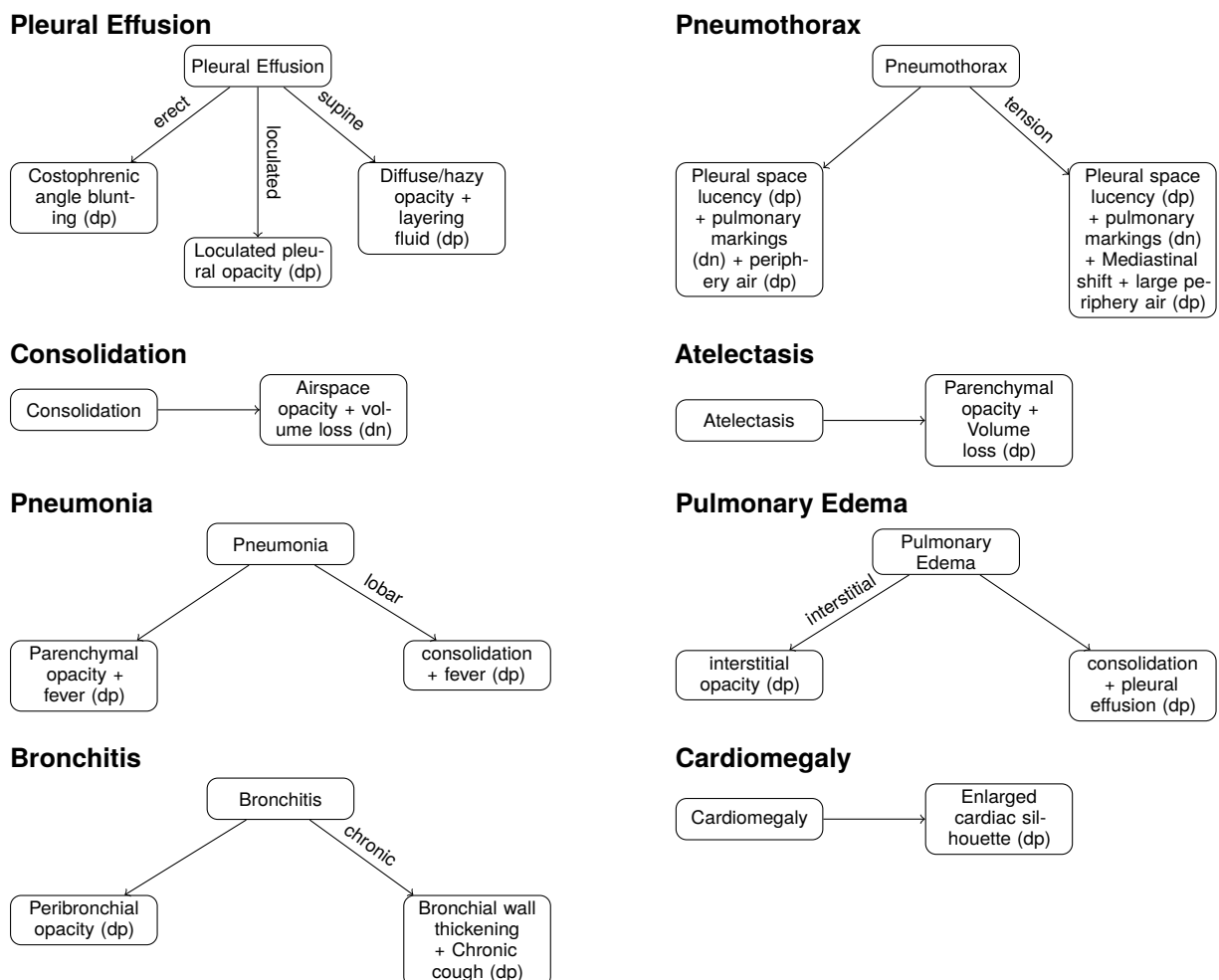

\begin{figure*}[tbp]
\centering
\scriptsize

\begin{minipage}[t]{0.45\linewidth}
\textbf{Congestive Heart Failure (CHF)}\\[0.3em]
\begin{tikzpicture}[node distance=0.7cm]
\node[dagnode] (chf) {CHF};

\node[dagnode, xshift=4cm, align=center, text width=3cm] (chf_combined) {Cardiomegaly (dp) +\\ Pulmonary edema (dp) +\\ Dyspnea / SOB (dp)};

\draw[->] (chf) -- (chf_combined);
\end{tikzpicture}
\end{minipage}
\hfill
\begin{minipage}[t]{0.45\linewidth}
\textbf{Emphysema}\\[0.3em]
\begin{tikzpicture}[node distance=0.7cm]
\node[dagnode] (emph) {Emphysema};

\node[dagnode, below=1cm of emph, xshift=-2.6cm] (luc) {Diffuse parenchymal lucency (dp)};
\node[dagnode, below=1cm of emph, xshift=+2.6cm] (vasc) {Diffuse parenchymal lucency (dp) + Decreased parenchymal destruction (dp)};

\draw[->] (emph) -- (luc); 
\draw[->] (emph) -- node[midway,sloped,above,font=\scriptsize]{severe} (vasc);
\end{tikzpicture}
\end{minipage}

\vspace{0.9em}

\begin{minipage}[t]{0.45\linewidth}
\textbf{COPD}\\[0.3em]
\begin{tikzpicture}[node distance=0.7cm]
\node[dagnode] (copd) {COPD};

\node[dagnode, xshift=+4cm, align=center, text width=3cm] (copd_combined) 
  {Hyperinflated lungs (dp) \\+ Emphysema (dp) +\\ Cough (dp)};

\draw[->] (copd) -- (copd_combined);
\end{tikzpicture}
\end{minipage}
\hfill
\begin{minipage}[t]{0.45\linewidth}
\textbf{Fracture}\\[0.3em]
\begin{tikzpicture}[node distance=0.7cm]
\node[dagnode] (fx) {Fracture};

\node[dagnode, below=1cm of fx, xshift=-2.6cm] (acute) {Bone disruption (dp)};
\node[dagnode, below=1cm of fx, xshift=+2.6cm] (old) {Bone callus formation (dp)};

\node[dagnode, below=2cm of fx, xshift=-1.3cm] (heal) {Bone deformity (dp)};
\node[dagnode, below=2cm of fx, xshift=+1.0cm] (comp) {Increased vertebral body opacity + height loss (dp)};

\draw[->] (fx) -- node[midway,sloped,above,font=\scriptsize]{acute} (acute);
\draw[->] (fx) -- node[midway,sloped,above,font=\scriptsize]{old} (old);
\draw[->] (fx) -- node[midway,sloped, above, font=\scriptsize]{healed/old} (heal);
\draw[->] (fx) -- node[midway,sloped, above,font=\scriptsize]{compression} (comp);
\end{tikzpicture}
\end{minipage}

\vspace{0.9em}

\begin{minipage}[]{0.45\linewidth}
\textbf{Tuberculosis}\\[0.3em]
\begin{tikzpicture}[node distance=0.7cm]
\node[dagnode] (tb) {Tuberculosis};

\node[dagnode, below=1cm of tb, xshift=-2.6cm] (active) {opacity  (dp)};
\node[dagnode, below=1.5cm of tb, xshift=+2.6cm] (chronic) {Calcified upper-lobe nodules + distortion (dp)};

\draw[->] (tb) -- node[midway,sloped,above,font=\scriptsize]{active} (active);
\draw[->] (tb) -- node[midway,sloped,above,font=\scriptsize]{old / inactive} (chronic);
\end{tikzpicture}
\end{minipage}
\hfill
\begin{minipage}[t]{0.45\linewidth}
\textbf{Lung Cancer}\\[0.3em]
\begin{tikzpicture}[node distance=0.7cm]
\node[dagnode] (lc) {Lung Cancer};

\node[dagnode, xshift=3cm] (mass) {Nodular or mass-like opacity (dp)};

\draw[->] (lc) -- (mass);
\end{tikzpicture}
\end{minipage}

\caption{
\textbf{Diagnostic pathway panels} (Part 2 of 2) illustrating exemplar pathways from the hand-crafted dictionary used by the pathway matching algorithm. Shown pathways correspond to CHF, Emphysema, COPD, Fracture, Tuberculosis, and Lung Cancer. CHF is modeled as a cardiopulmonary congestion cluster, where cardiomegaly, pulmonary edema, and dyspnea co-occur during exacerbation. Emphysema encodes a severity axis from diffuse parenchymal lucency to severe parenchymal destruction with decreased peripheral vascularity. COPD is represented as a chronic obstructive cluster combining hyperinflated lungs, emphysematous change, and chronic cough. Fracture pathways separate acute cortical breaks, old or healed deformity, and vertebral compression patterns. Tuberculosis includes an explicit disease-state axis, contrasting active parenchymal opacity with old or inactive upper-lobe scarring that shows calcified nodules and architectural distortion. Lung Cancer is represented by a focal nodular or mass-like opacity that may reflect primary or metastatic disease.}
\label{fig:pathway_panels_part2_final6}
\end{figure*}
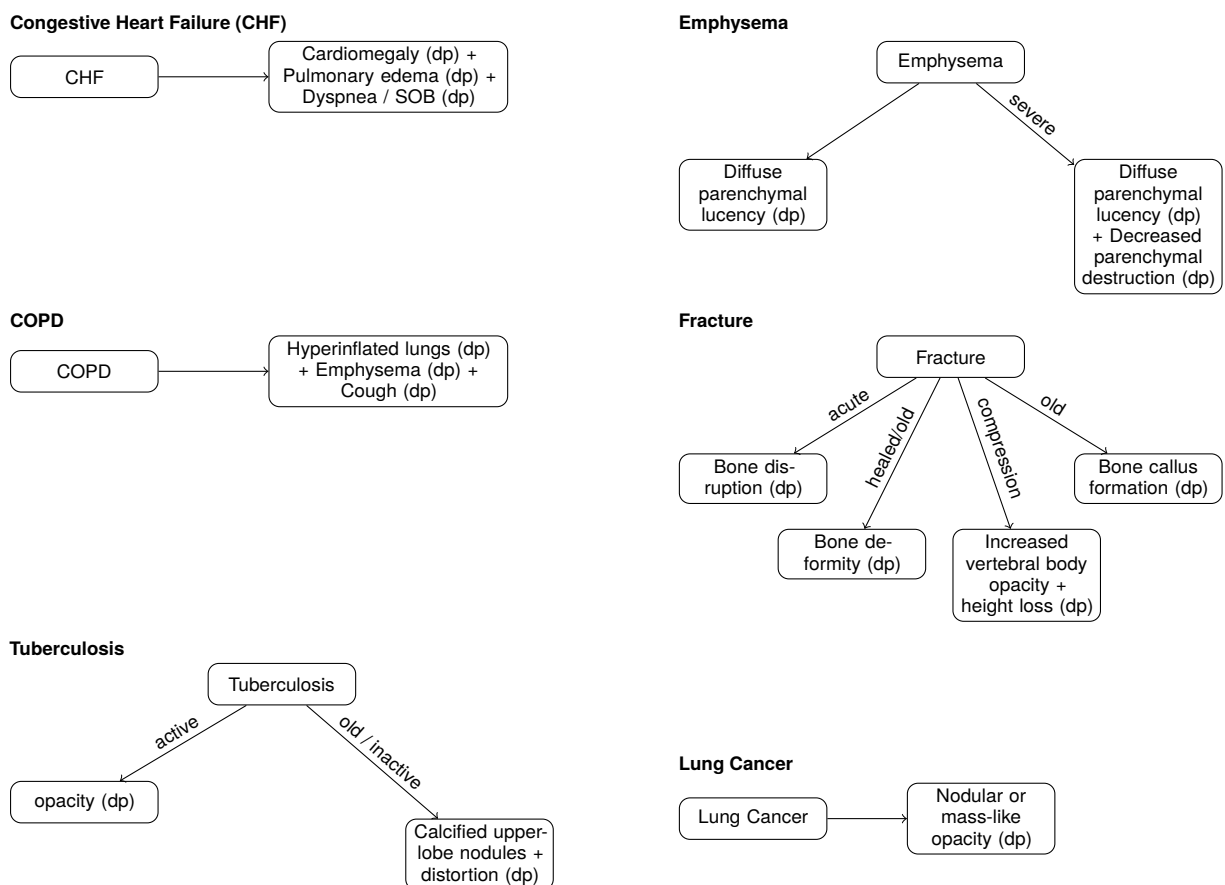

\subsection{Cascading Expansion Framework}
\label{appendix:expansion_framework}

\subsubsection{Blacklist for Finding Deduplication}
\label{app:blacklist}

We merge near-duplicate findings into a single canonical form using cosine similarity from a clinical embedding model (e.g., BioLord~\citep{remy2023biolord}) with a threshold of $0.9$. Even with this high threshold, purely lexical similarity can still spuriously merge clinically distinct statements that differ along critical axes (e.g., \emph{\textbf{left} lower lobe new consolidation} vs. \emph{\textbf{right} lower lobe new consolidation}).
To avoid false merges of distinct concepts, we apply an explicit blacklist with two layers: (i) exact-pair exclusions and (ii) pattern-level rules. 
We use a manually curated blacklist of mutually incompatible pairs, which is invoked when two candidates belong to the same study and to the same coarse anatomic region but differ along key discriminative axes such as laterality, lobar/segmental location, diagnostic status, or pathophysiologic mechanism. 
A candidate pair is merged only if it passes the similarity threshold and does not match any blacklist entry; otherwise both findings are preserved. 
Representative blacklist entries and patterns are summarized in Table~\ref{tab:blacklist_pairs}. 

\begin{table*}[htbp]
\centering
\small
\begin{tabular}{ll}
\toprule
\textbf{Category} & \textbf{Non-merge pairs (substring-level)} \\
\midrule
Laterality and anatomical direction &
\begin{tabular}[t]{@{}l@{}}
(left, right), (left-sided, right-sided), (left-sided, right), (left, right-sided), \\
(upper, lower), (mid, lower), (upper, mid), (upper, middle), (middle, lower), \\
(anterior, posterior), (anterior, lateral), (posterior, lateral), \\
(superior, inferior), (apical, basal), (central, peripheral), \\
(proximal, distal), (medial, lateral), (ventral, dorsal)
\end{tabular}
\\[6pt]
Cardiopulmonary contextual conflicts &
(cardio, pulmonary), (mediastinal, pleural), (pericardial, pleural)
\\[4pt]
Disease mechanism divergence &
\begin{tabular}[t]{@{}l@{}}
(effusion, pneumothorax), (effusion, edema), (effusion, atelectasis), \\
(effusion, consolidation), (effusion, pneumonia), \\
(consolidation, opacification), (consolidation, atelectasis), \\
(consolidation, edema), (atelectasis, pneumonia), \\
(atelectasis, aeration), (atelectasis, opacity), (pneumonia, edema)
\end{tabular}
\\[6pt]
Mass vs. airspace process &
(mass, consolidation), (mass, opacity), (mass, atelectasis)
\\
\bottomrule
\end{tabular}
\caption{\textbf{Manually curated blacklist} of substring pairs that prevent merges during high-precision deduplication.}
\label{tab:blacklist_pairs}
\normalsize
\end{table*}

\subsubsection{Details of Pathway Matching}
\label{appendix:ExpansionRules}
Algorithm~\ref{alg:match} describes the pathway matching procedure, which aligns each input tuple \((\textit{finding}, \textit{loc}, \textit{attr}, \textit{view})\) to at most one pathway variant from Table \ref{tab:pathway_summary} and Figures \ref{fig:pathway_panels_part1_final8} and \ref{fig:pathway_panels_part2_final6}. The full code can be found on our \href{https://github.com/prabaey/lunguage_uncertainty/blob/main/run_pathway/dx_pathway.py}{Github repository}. 

For each row from \textsc{Lunguage}, we first apply \texttt{NormalizeInput} to obtain a normalized tuple \((f',\ell',a',v')\), mapping raw report phrases to the \textsc{Lunguage} vocabulary (e.g., ``RUL'' \(\rightarrow\) ``right upper lung,'' ``TB'' \(\rightarrow\) ``tuberculosis'') and mapping view metadata to a standardized \texttt{view\_information} string (AP, PA, LATERAL, erect, supine). 
In parallel, we apply \texttt{NormalizeDict} to the pathway dictionary \(\mathcal{D}\), yielding \(\mathcal{D}'\) with normalized findings, locations, attributes, and views for every pathway entry.

Given \((f',\ell',a',v')\) and \(\mathcal{D}'\), the \texttt{BestPathwayMatch} routine proceeds in three compatibility steps.
First, it performs a \emph{finding-level lexical match}: we restrict candidates to those pathways whose normalized finding term is identical to \(f'\). 
This ensures that, for example, tuples with finding ``fracture'' only compete among fracture-related pathways, and tuples with finding ``pleural effusion'' only among effusion pathways.

Second, we enforce \emph{location and attribute compatibility}. 
For example, location compatibility distinguishes skeletal fractures from device fractures even when the normalized finding f’ is “fracture”.
Attribute terms then select more specific variants: ``loculated'' triggers the loculated-effusion pathway, ``tension'' the tension-pneumothorax pathway, ``compression'' the compression-fracture pathway, and acuity modifiers such as ``acute,'' ``old,'' or ``healed'' select the corresponding fracture branch.

Finally, we check \emph{view compatibility} using \texttt{view\_information}, retaining only variants that are observable under the given view (e.g., erect versus supine presentations of pleural effusion, as discussed earlier). 
By construction, these compatibility rules are defined so that at most one pathway variant remains for any normalized input tuple. 
If a unique compatible variant exists, \texttt{BestPathwayMatch} returns it as \(c^\star\); otherwise it returns \textit{none}. 
\begin{algorithm}[htbp]
\caption{Pathway Matching}
\label{alg:match}
\scriptsize
\begin{algorithmic}[1]
\newcommand{\mycomment}[1]{\textcolor{gray}{\scriptsize\textit{#1}}}

\Input $(f,\ell,a,v)$ from \textsc{Lunguage}, pathway dict $\mathcal{D}$
\mycomment{// $f$: finding, $\ell$: location, $a$: attributes, $v$: view}
\Output $c^\star$ or \textit{none}

\Function{SelectVariant}{$f,\ell,a,v,\mathcal{D}$}
  \State $(f',\ell',a',v') \leftarrow \mathrm{NormalizeInput}(f,\ell,a,v)$
         \mycomment{// normalize row-level finding, location, attributes, and view}
  \State $\mathcal{D}' \leftarrow \mathrm{NormalizeDict}(\mathcal{D})$
         \mycomment{// normalize findings, locations, attributes, and views in all pathways}
  \State $c^\star \leftarrow \mathrm{BestPathwayMatch}(\mathcal{D}', f',\ell',a',v')$
  \State \Return $c^\star$
\EndFunction

\Function{BestPathwayMatch}{$\mathcal{D}', f',\ell',a',v'$}
  \State $C \leftarrow \{\,c \in \mathcal{D}' : \mathrm{Find}(c) = f'\,\}$%
         \mycomment{// finding-level lexical match}
  \If{$C = \emptyset$} \State \Return \textit{none} \EndIf
  \State $C \leftarrow \{\,c \in C : \mathrm{LocOK}(c,\ell') \land \mathrm{AttrOK}(c,a') \land \mathrm{ViewOK}(c,v')\,\}$%
         \mycomment{// joint compatibility on location, attributes, and view}
  \If{$|C| = 1$}
    \State \Return \text{the sole element of } $C$
  \Else
    \State \Return \textit{none}%
         \mycomment{// ambiguous or incompatible (should not occur since pathways are independent)}
  \EndIf
\EndFunction
\end{algorithmic}
\end{algorithm}

\subsubsection{Status Conflict Resolution}
\label{app:conflict_analysis}
During pathway expansion we observed status conflicts for the same finding, and we also identified inconsistencies within original reports. In Table \ref{tab:conflict_crosstab}, we categorize conflicts by \emph{source} and \emph{type} and count their occurrence in \textsc{Lunguage}. Sources are \texttt{original\_vs\_expansion} (an expanded node contradicts an explicitly stated node), \texttt{original\_vs\_original} (two original nodes disagree), and \texttt{expansion\_vs\_expansion} (two expanded nodes disagree). Types are \texttt{polarity} (dp vs.\ dn), \texttt{certainty\_positive} (dp vs.\ tp), \texttt{certainty\_negative} (dn vs.\ tn), \texttt{duplicate\_pos} (tn vs. tn) and \texttt{duplicate\_neg} (tp vs. tp).

Conflict detection operates at the granularity of (\texttt{entity},\allowbreak \texttt{location}) within each report with the same \texttt{study\_id}. Across the expanded \textsc{Lunguage} dataset (19{,}216 rows), expansion-related conflicts were rare overall, totaling 616 cases (\textbf{3.2\%}): 165 (\textbf{0.9\%}) \texttt{original\_vs\_expansion} and 451 (\textbf{2.3\%}) \texttt{expansion\_vs\_expansion}. Within \texttt{original\_vs\_expansion} (n=165), conflicts due to polarity of definitive findings dominated, with \texttt{polarity} accounting for 107 (64.8\%) conflicts, \texttt{certainty\_pos} for 34 (20.6\%), and \texttt{certainty\_neg} for 24 (14.5\%). Within \texttt{expansion\_vs\_expansion} (n=451), conflicts were dominated by duplicates of positively tentative findings: \texttt{polarity} accounted for 103 (22.8\%) conflicts, \texttt{certainty\_pos} for 133 (29.5\%), \texttt{certainty\_neg} for 16 (3.5\%), \texttt{duplicate\_pos} for 181 (40.1\%), and \texttt{duplicate\_neg} for 18 (4.0\%). In addition, some inconsistencies were already present in the original reports: \texttt{original\_vs\_original} totaled 158 cases (\textbf{0.8\%}). These conflicts were dominated by positive certainty conflicts: \texttt{polarity} accounted for 10 (6.3\%) conflicts, \texttt{certainty\_pos} for 74 (46.8\%), \texttt{certainty\_neg} for 2 (1.3\%), \texttt{duplicate\_pos} for 70 (44.3\%), and \texttt{duplicate\_neg} for 2 (1.3\%). Overall conflicts across all sources summed to 774 (\textbf{4.0\%} of the initial expanded \textsc{Lunguage} dataset).

Resolution follows a deterministic policy: (1) if a group contains both original and expansion rows, retain the originals and discard the expansions (treat the report text as the clinical source of truth); (2) if the remaining rows show a pure polarity clash with only \texttt{dp} and \texttt{dn}, drop the group; (3) otherwise select the row with the highest \texttt{prob}. When multiple rows tie on \texttt{prob}, break ties by a status priority that reflects clinical certainty, \texttt{dp} $>$ \texttt{tp} $>$ \texttt{tn} $>$ \texttt{dn}. After resolution, \textbf{18{,}810} rows remained (\textbf{97.9\%} of 19{,}216), implying \textbf{406} removals (\textbf{2.1\%}); all remaining inconsistencies were eliminated, preserving logical consistency and clinical validity. Algorithm~\ref{alg:resolve_general} details the procedure.

\begin{table}[htbp]
\centering
\resizebox{\columnwidth}{!}{
\begin{tabular}{llccc}
\toprule
\textbf{Conflict} & \textbf{Type} & exp $\leftrightarrow$ exp & ori $\leftrightarrow$ exp & ori $\leftrightarrow$ ori \\
\midrule
dn $\leftrightarrow$ tn & \texttt{certainty\_neg} & 16  & 0   & 2  \\
dp $\leftrightarrow$ tp & \texttt{certainty\_pos} & 133 & 34  & 74 \\
tn $\leftrightarrow$ tn & \texttt{duplicate\_neg} & 18  & 0   & 2  \\
tp $\leftrightarrow$ tp & \texttt{duplicate\_pos} & 181 & 24  & 70 \\
dp $\leftrightarrow$ dn & \texttt{polarity} & 103 & 107 & 10 \\
\midrule
& & 451 & 165 & 158 \\
\bottomrule
\end{tabular}
}
\caption{\textbf{Pathway expansion conflicts} by source and type. \textit{ori} and \textit{exp} denote original and expansion. 
}
\label{tab:conflict_crosstab}
\end{table}

\newcommand{\mycomment}[1]{\textcolor{gray}{\scriptsize\textit{#1}}}

\begin{algorithm}[t]
\caption{Conflict Resolution}
\label{alg:resolve_general}
\scriptsize
\begin{algorithmic}[1]
\Input Expanded dataset $D$ with columns:
\texttt{study\_id}, \texttt{entity}, \texttt{location},
\texttt{status} $\in\{\texttt{dp},\texttt{tp},\texttt{tn},\texttt{dn}\}$, \texttt{prob},
\texttt{source} $\in\{\texttt{original},\texttt{expansion}\}$
\Output \texttt{resolved\_df}

\State keys $\gets$ [\texttt{study\_id}, \texttt{entity}, \texttt{location}]
\State \texttt{RES} $\gets [\,]$
\ForAll{group $G \subset D$ by \texttt{keys}}
  \If{$|G|<2$} \textbf{continue} \EndIf 

  \State $\texttt{orig} \gets \{\, r \in G \mid r.\texttt{source}=\texttt{original} \,\}$
  \State $\texttt{expd} \gets \{\, r \in G \mid r.\texttt{source}=\texttt{expansion} \,\}$

  \mycomment{--- Case 1: original vs. expansion ---}
  \If{$\texttt{orig}\neq\varnothing$ \textbf{and} $\texttt{expd}\neq\varnothing$}
  
    \mycomment{Rule1: keep originals only; discard expansions}

    \State append all rows in \texttt{orig} to \texttt{RES}
    \State \textbf{continue}
  \EndIf

  \mycomment{--- Case 2: only originals or only expansions ---}
  \State $R \gets \texttt{orig}$ if $\texttt{orig}\neq\varnothing$ else $\texttt{expd}$
  \State $has\_dp \gets (\exists r\in R:\ r.\texttt{status}=\texttt{dp})$
  \State $has\_dn \gets (\exists r\in R:\ r.\texttt{status}=\texttt{dn})$
  \State $has\_tp \gets (\exists r\in R:\ r.\texttt{status}=\texttt{tp})$
  \State $has\_tn \gets (\exists r\in R:\ r.\texttt{status}=\texttt{tn})$

  \mycomment{Rule2: drop if pure polarity clash}
  
  \If{$has\_dp$ \textbf{and} $has\_dn$ \textbf{and} \textbf{not}($has\_tp$ \textbf{or} $has\_tn$)}
     \State \textbf{continue}
  \EndIf

  \mycomment{Rule 3: pick highest probability; dp > tp > tn > dn}
  \State $p^\ast \gets \max\{\, r.\texttt{prob} : r \in R \,\}$
  \If{$\exists r\in R : r.\texttt{prob}=p^\ast \land r.\texttt{status}=\texttt{dp}$}
     \State append any such $r$ to \texttt{RES}
  \ElsIf{$\exists r\in R : r.\texttt{prob}=p^\ast \land r.\texttt{status}=\texttt{tp}$}
     \State append any such $r$ to \texttt{RES}
  \ElsIf{$\exists r\in R : r.\texttt{prob}=p^\ast \land r.\texttt{status}=\texttt{tn}$}
     \State append any such $r$ to \texttt{RES}
  \Else
     \State append any $r\in R$ with $r.\texttt{prob}=p^\ast$ to \texttt{RES}
  \EndIf
\EndFor

\State \texttt{resolved\_df} $\gets \mathrm{Concat}(\texttt{RES})$
\State \Return \texttt{resolved\_df}
\end{algorithmic}
\end{algorithm}

\begin{table*}[p]
\centering
\scriptsize
\begin{tabularx}{\textwidth}{
  >{\raggedright\arraybackslash}p{0.17\textwidth}
  >{\raggedright\arraybackslash}p{0.28\textwidth}
  >{\raggedright\arraybackslash}X
}
\toprule
\textbf{Diagnosis} & \textbf{Specific diagnosis} & \textbf{Diagnostic pathways} \\
\midrule

Pleural Effusion & Non-loculated pleural effusion \textit{(one of nine pathways)} & \texttt{view: ap, pa, lateral > ent: blunting > status: dp > loc: costophrenic angle \&\& ent: opacity > status: dp > loc: pleural space > attr: hazy, diffuse}\\
& Loculated pleural effusion & \texttt{view: ap, pa, lateral > ent: opacity > status: dp > loc: pleural space > attr: loculated} \\
\addlinespace[2pt]

Pneumothorax & Pneumothorax & \texttt{view: ap, pa, lateral > ent: lucency > status: dp > loc: pleural space \&\& ent: marking > status: dn > loc: pulmonary \&\& ent: air > status: dp > loc: lung periphery}\\
& Tension pneumothorax & \texttt{Pneumothorax \&\& ent: shift > status: dp > loc: mediastinal \&\& attr: large amount}\\
\addlinespace[2pt]

Consolidation & &
\texttt{view: ap, pa, lateral > ent: opacity > status: dp > loc: airspace \&\& ent: volume loss > status: dn} \\
\addlinespace[2pt]

Atelectasis & &
\texttt{view: ap, pa, lateral > ent: opacity > status: dp \&\& ent: volume loss > status: dp} \\
\addlinespace[2pt]

Pneumonia & Pneumonia  & \texttt{view: ap, pa, lateral > ent: opacity > status: dp \&\& ent: fever > status: dp}\\
&  Aspiration pneumonia & \texttt{Pneumonia > loc: lung > attr: dependent portion}\\
& Lobar pneumonia & \texttt{view: ap, pa, lateral > ent: consolidation > status: dp \&\& ent: fever > status: dp} \\
\addlinespace[2pt]

Pulmonary Edema & Interstitial pulmonary edema & \texttt{view: ap, pa, lateral > ent: opacity > status: dp > loc: interstitial}\\
& Alveolar pulmonary edema & \texttt{view: ap, pa, lateral > ent: consolidation > status: dp \&\& ent: pleural effusion > status: dp} \\
\addlinespace[2pt]

Bronchitis & Bronchitis & \texttt{view: ap, pa, lateral > ent: opacity > status: dp > loc: peribronchial}\\
& Chronic Bronchitis & \texttt{view: ap, pa, lateral > ent: thickening > status: dp > loc: bronchial wall \&\& ent: cough > status: dp} \\
\addlinespace[2pt]

Cardiomegaly &  &
\texttt{view: ap, pa > ent: heart size > status: dp > loc: cardiothoracic > attr: increased} \\
\addlinespace[2pt]

CHF &  &
\texttt{view: ap, pa, lateral > ent: cardiomegaly > status: dp \&\& ent: pulmonary edema > status: dp \&\& ent: dyspnea > status: dp} \\
\addlinespace[2pt]

Emphysema & Emphysema & \texttt{view: ap, pa, lateral > ent: lucency > status: dp > loc: lung parenchyma > attr: diffuse}\\
& Severe emphysema & \texttt{Emphysema \&\& ent: pulmonary vascularity > status: dp > attr: decreased \&\& ent: destruction > status: dp > loc: lung parenchyma} \\
\addlinespace[2pt]

COPD & &
\texttt{view: ap, pa, lateral > ent: hyperinflation > status: dp > loc: lungs \&\& ent: emphysema > status: dp \&\& ent: cough > status: dp} \\
\addlinespace[2pt]

Fracture & Acute fracture &
\texttt{view: ap, pa, lateral > ent: disruption > status: dp > loc: bone}\\
& Chronic/Old fracture & \texttt{view: ap, pa, lateral > ent: callus formation > status: dp > loc: bone}\\
& Healed fracture & \texttt{view: ap, pa, lateral > ent: deformity > status: dp > loc: bone}\\
& Spinal/Compression fracture & \texttt{view: ap, pa, lateral > ent: opacity > status: dp > loc: vertebral body > attr: increased \&\& ent: loss of height > status: dp} \\
\addlinespace[2pt]

Tuberculosis & Active tuberculosis & \texttt{view: ap, pa, lateral > ent: opacity > status: dp}\\
& Chronic/Non-active tuberculosis &\texttt{view: ap, pa, lateral > ent: nodules > status: dp > loc: bilateral upper lung > attr: calcified \&\& ent: architectural distortion > status: dp > loc: bilateral upper lung} \\
\addlinespace[2pt]

Lung Cancer & &
\texttt{view: ap, pa, lateral > ent: opacity > status: dp > attr: nodular} \\
\bottomrule
\end{tabularx}
\caption{\textbf{Diagnostic pathways.} \texttt{view} denotes projection or patient orientation; \texttt{ent}, \texttt{status}, \texttt{loc}, and \texttt{attr} indicate the entity, status (\texttt{dp} is definitive positive, \texttt{dn} is definitive negative), anatomical location, and attributes (e.g., morphology, distribution, measurement). Pathways are ordered sequences joined by ``\texttt{>}'', with required co-occurrence marked by ``\texttt{\&\&}''. We display one representative pathway out of nine defined for pleural effusion and, in total, 33 pathways spanning 14 diagnoses.}
\label{tab:pathway_summary}
\end{table*}

\end{document}